\documentclass{article}

\usepackage[preprint]{neurips_2024}

\usepackage[utf8]{inputenc} 
\usepackage[T1]{fontenc}    
\usepackage{hyperref}       
\usepackage{url}            
\usepackage{booktabs}       
\usepackage{amsfonts}       
\usepackage{nicefrac}       
\usepackage{microtype}      
\usepackage{xcolor}         

\usepackage{amsmath}
\usepackage{multirow}
\usepackage{ragged2e}
\usepackage{tabularx}
\usepackage{graphicx}
\usepackage{float}
\usepackage{subcaption}
\usepackage{soul}
\usepackage[most]{tcolorbox}
\usepackage{relsize}
\usepackage[font=small,labelfont=bf]{caption}
\usepackage{wrapfig}
\usepackage[lined,ruled,commentsnumbered]{algorithm2e}
\usepackage{algpseudocode}

\definecolor{ForestGreen}{RGB}{34,139,34}

\newcommand{\smallforestgreen}[1]{{\smaller\smaller\textcolor{ForestGreen}{#1}}}
\newcommand{\smallforestred}[1]{{\smaller\smaller\textcolor{red}{#1}}}

\def\panguR{Pangu-R-38B}
\def\ours{TrimR}

\title{TrimR: Verifier-based Training-Free Thinking Compression for Efficient Test-Time Scaling}
\vspace{-0.5cm}
\author{%
  Weizhe Lin $^1$$\textsuperscript{\textdagger}$
  \quad
  Xing Li $^2$$\textsuperscript{\textdagger}$
  \quad
  Zhiyuan Yang $^2$ \textsuperscript{\textdagger}
  \quad
  Xiaojin Fu $^2$ \textsuperscript{\textdagger}
  \quad
  Hui-Ling Zhen $^2$ \\
  \quad
  \textbf{Yaoyuan Wang} $^1$
  \quad
  \textbf{Xianzhi Yu} $^2$
  \quad
  \textbf{Wulong Liu} $^2$
  \quad
  \textbf{Xiaosong Li} $^1$
  \quad
  \textbf{Mingxuan Yuan} $^2$
\\
  $^1$ Huawei Advanced Computing and Storage Lab
  $^2$ Huawei Noah's Ark Lab \\
  \textit{\{linweizhe1, li.xing2, YangZhiyuan123, fuxiaojin, zhenhuiling2, wangyaoyuan1,}\\ \textit{yuxianzhi, liuwulong, lixiaosong20, yuan.mingxuan\}@huawei.com}
}

\begin{document}

\maketitle

\def\thefootnote{\textsuperscript{\textdagger}}\footnotetext{Equal contribution.}\def\thefootnote{\arabic{footnote}}

\vspace{-0.5cm}
\begin{abstract}
\vspace{-0.3cm}
Large Reasoning Models (LRMs) demonstrate exceptional capability in tackling complex mathematical, logical, and coding tasks by leveraging extended Chain-of-Thought (CoT) reasoning. Test-time scaling methods—such as prolonging CoT with explicit token-level exploration—can push LRMs’ accuracy boundaries, but they incur significant decoding overhead. 
A key inefficiency source is LRMs often generate redundant thinking CoTs, which demonstrate clear structured overthinking and underthinking patterns.
Inspired by human cognitive reasoning processes and numerical optimization theories, we propose \textbf{\textit{TrimR}}, a \textbf{verifier-based, training-free, efficient framework for dynamic CoT compression to \underline{trim} \underline{r}easoning and enhance test-time scaling}, explicitly tailored for production-level deployment. Our method employs a lightweight, pretrained, instruction-tuned verifier to detect and truncate redundant intermediate thoughts of LRMs without any LRM or verifier fine-tuning. We present both the core algorithm and asynchronous online system engineered for high-throughput industrial applications.
Empirical evaluations on \textbf{Ascend NPUs and vLLM} show that our framework delivers substantial gains in inference efficiency under large-batch workloads. In particular, on the four MATH500, AIME24/25, and GPQA benchmarks, \textbf{the reasoning runtime of Pangu Pro MoE, \panguR{}, QwQ-32B, and DeepSeek-R1-Distill-Qwen-32B is improved by up to 70\% with negligible impact on accuracy}.
\end{abstract}
\begin{figure}[h]
    \vspace{-0.7cm}
  \centering
  \includegraphics[width=0.9\linewidth]{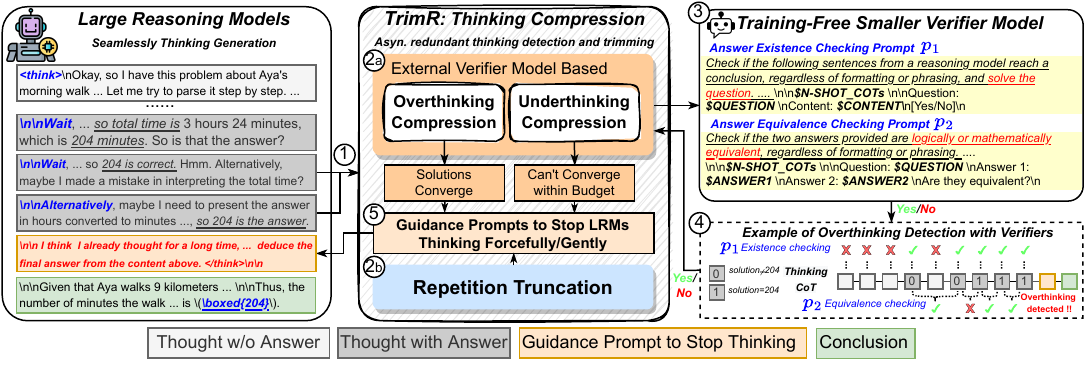}
  \caption{The TrimR thinking compression framework for trimming reasoning.
  (1) CoTs are divided into sub-thoughts with reflection tokens in Sec. \ref{sec:thought_detection}; (2a) Over/underthinking detection through answer convergence analysis in Sec. \ref{sec:overthinking_compression}, \ref{sec:underthinking_compression} (2b) repetition truncation in Sec. \ref{sec:repetition_truncation};
  (3) By simplifying redundant detection as the binary answer existence and equivalence classification with prompts $p_1$ and $p_2$ in Appendix \ref{sec:detailed_prompts}, 7B instruction models replace PRMs/ORMs, avoiding instability from full-sequence scoring;
  (4) The verifier applies $p_1$ to assess conclusion completeness in individual thoughts and $p_2$ to identify overthinking when consecutive thoughts yield identical answers. Early termination is triggered with Algorithm \ref{alg:overthinking_compression}, \ref{alg:underthinking_compression};
  (5) Thinking termination prompts for LRMs are generated based on verifier decisions to halt redundant reasoning gently or forcefully in Sec. \ref{sec:guidance_prompt_lrm}.}
  \label{fig:thinking_compression_framework}
\end{figure}

\vspace{-0.3cm}
\section{Introduction}
\vspace{-0.1cm}

Large Reasoning Models (LRMs) such as OpenAI o1~\cite{o1openai}, DeepSeek R1~\cite{guo2025deepseek}, Qwen QwQ~\cite{qwq32b}, and Pangu Pro MoE~\cite{tang2025pangupromoe} achieve expert‐level performance by decomposing problems into manageable subproblems, conducting step‐by‐step analysis, detecting and correcting errors, and exploring alternative solutions. However, this extended “thinking” incurs substantial decoding overhead and inference cost, hindering deployment in resource‐constrained settings~\cite{guo2025deepseek}.

\begin{figure}
    \centering
    \begin{subfigure}{0.245\columnwidth}
        \centering
        \includegraphics[width=\columnwidth]{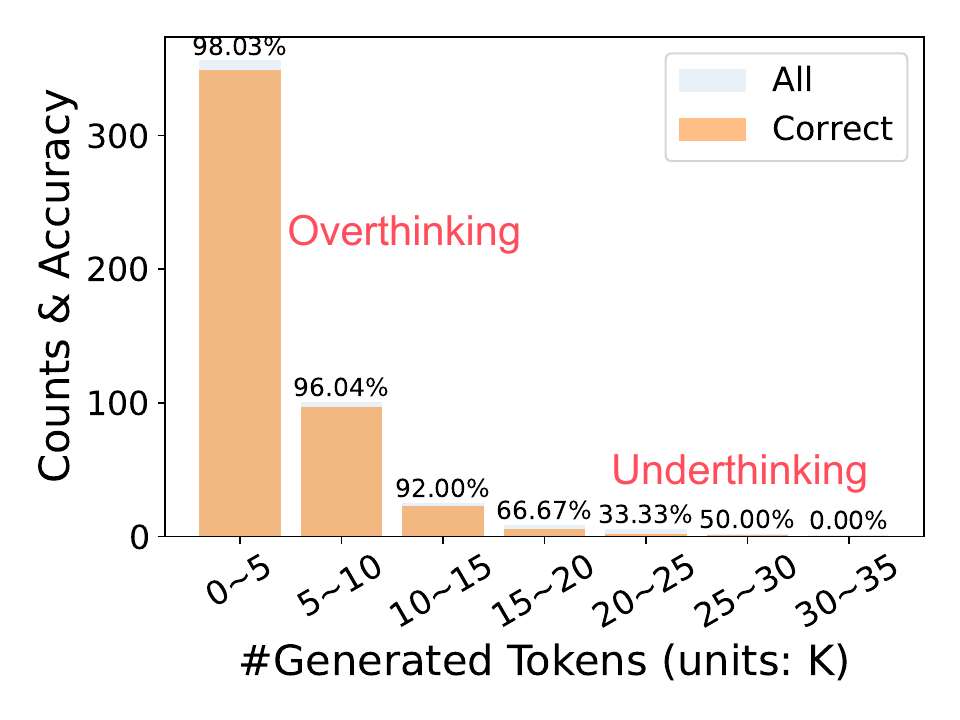}
        \vspace{-1.5em}
        \caption{MATH500 Accuracy}
        \label{fig:count_correctness_qwq_32b_math500}
    \end{subfigure}
    \begin{subfigure}{0.245\columnwidth}
        \centering
        \includegraphics[width=\columnwidth]{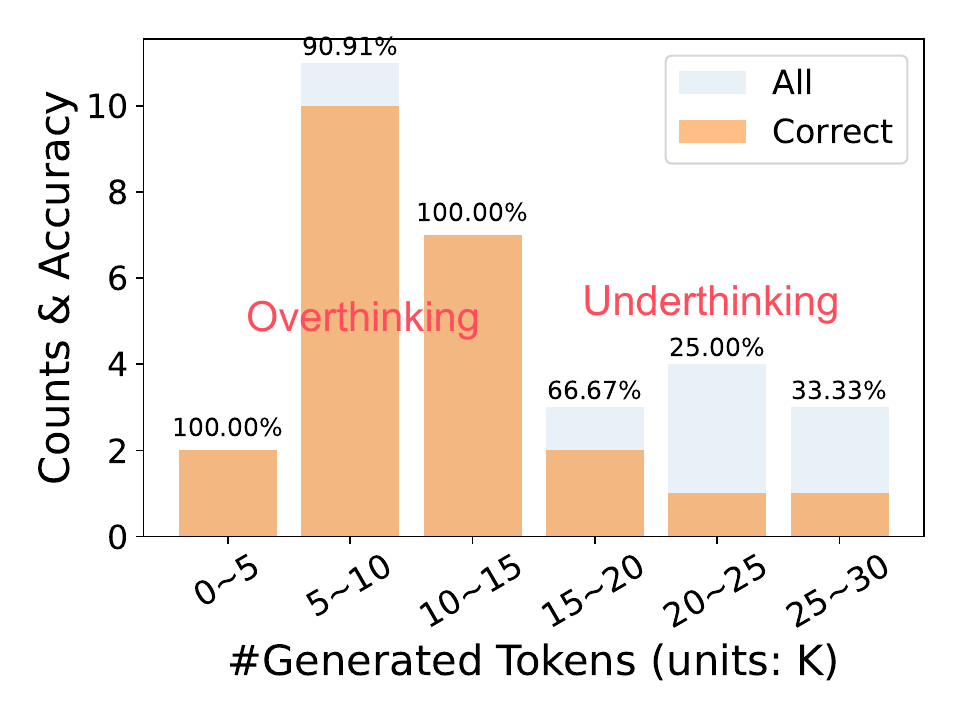}
        \vspace{-1.5em}
        \caption{AIME24 Accuracy}
        \label{fig:count_correctness_qwq_32b_aime24}
    \end{subfigure}
    \begin{subfigure}{0.245\columnwidth}
        \centering
        \includegraphics[width=\columnwidth]{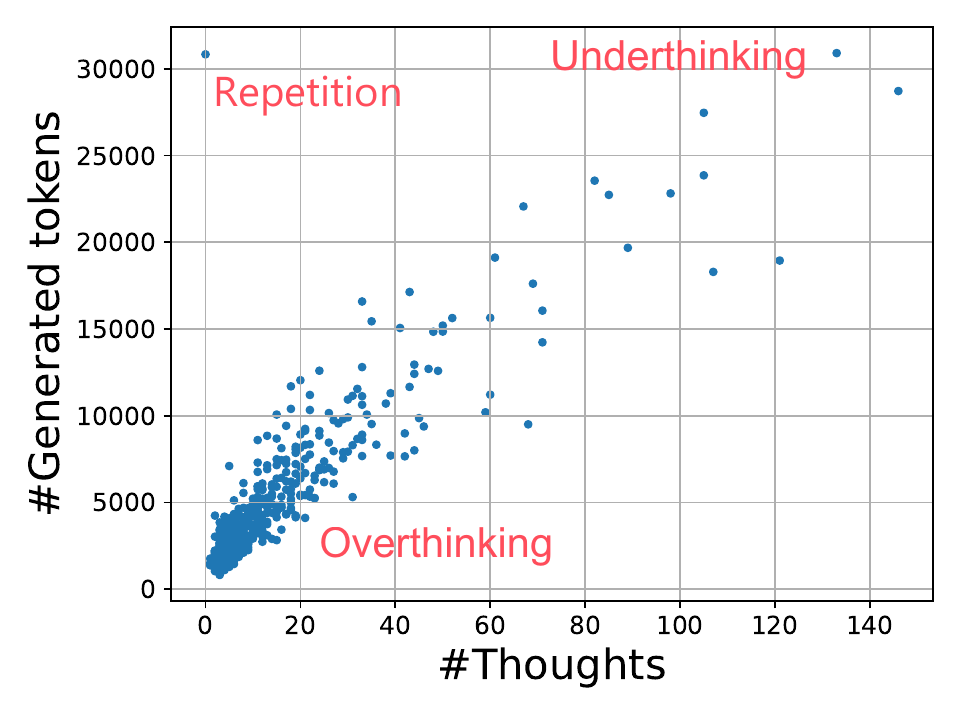}
        \vspace{-1.5em}
        \caption{MATH500 \#Thoughts}
        \label{fig:num_thought_tokens_qwq_32b_math500}
    \end{subfigure}
    \begin{subfigure}{0.245\columnwidth}
        \centering
        \includegraphics[width=\columnwidth]{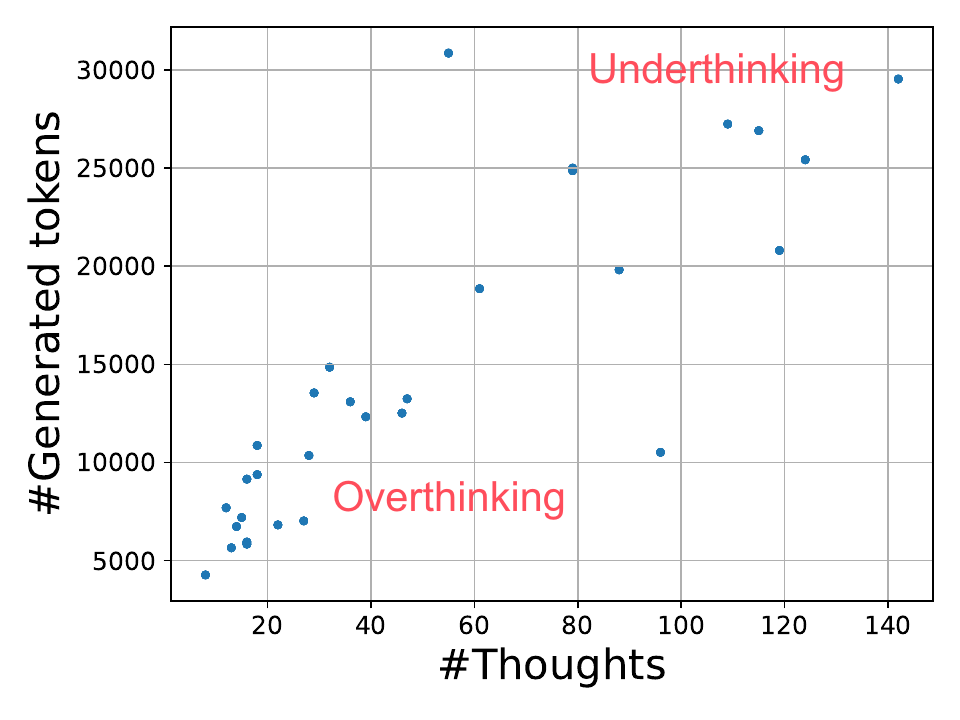}
        \vspace{-1.5em}
        \caption{AIME24 \#Thoughts}
        \label{fig:num_thought_tokens_qwq_32b_aime24}
    \end{subfigure}
    \vspace{-0.6cm}
    \caption{Correlation between occurrence, accuracy, and the number of generated thoughts/tokens of QwQ-32B on MATH500~\cite{hendrycks2021math} and AIME24~\cite{aime24}. 
    For hard questions, QwQ can spin out up to 140 lengthy yet incorrect thoughts—indicating a need for underthinking compression.
    For easier questions, QwQ delivers brief, highly accurate answers—yet there’s still room to make them even more concise by overthinking compression.
    }
    \label{fig:count_correctness_qwq}
    \vspace{-0.4cm}
\end{figure}

Benchmarking on AIME24~\cite{aime24} and MATH500~\cite{hendrycks2021math} reveals two key inefficiencies: \textbf{overthinking} and \textbf{underthinking}~\cite{marjanovic2025deepseekThoughtPatterns}. \textbf{Overthinking} manifests as redundant verification of already‐correct steps—often signaled by reflection tokens such as “Wait” or “Alternatively”—which increases output length without accuracy gains (as in Figs.~\ref{fig:thinking_compression_framework},~\ref{fig:count_correctness_qwq}). \textbf{Underthinking} occurs on harder problems when the model oscillates among incomplete reasoning chains and fails to converge, producing lengthy yet inaccurate responses (Fig.~\ref{fig:count_correctness_qwq}). Representative examples are provided in Appendix~\ref{sec:overthinking_underthinking_examples}.

The extended Chain‐of‐Thought (CoT) reasoning in LRMs poses deployment challenges: decoding lengths vary widely and runtime scales superlinearly with sequence length. Training-based methods such as \cite{yan2025inftythink,chen2024tencent_simpo,munkhbat2025self,yu2024distilling} reduce token generation in LRMs but impose heavy training and computational costs on large models and may degrade their general capabilities. In contrast, training-free techniques integrate seamlessly and preserve original behavior: TokenBudget~\cite{han2024token} dynamically adjusts token budgets via prompting, and Chain of Draft~\cite{xu2025chain} uses concise instructions to shorten output, yet both require invasive inference-time modifications.

We thus propose \textbf{\textit{\ours}}, a \textbf{verifier-based, training-free approach for online thinking compression} while maintaining reasoning performance. We detect overthinking and underthinking in \textbf{intermediate thought answers} with a lightweight verifier and trigger prompt-based LRM thinking early termination upon redundancy detection. By simply checking answer existence and equivalence in brief thought segments, our method leverages compact verifier models instead of complex Process Reward Models (PRMs) or Outcome Reward Models (ORMs)~\cite{lightman2024lets}. The non-invasive early termination preserves original LRM capabilities. Finally, we present an online asynchronous system collaborating LRMs and small verifiers to support industrial-scale workloads.

\textbf{Notations.}
Given an input $X$, the LRM $\Pi$ generates a response $Y = \Pi(X)$. We denote $\mathbf{y}_{<t}=[y_1, y_2, ...,y_{t-1}]$ as the previously generated tokens. Each time $y_t=p_{\Pi}(X, \mathbf{y}_{<t})$. With a slight abuse of notation, let $\Pi(X, \mathbf{y}_{<t})$ denote full response including $\mathbf{y}_{<t}$. 

\textbf{Theoretical Foundation.}
Our method unifies human cognitive heuristics—overthinking and underthinking as confidence‐threshold and diminishing‐returns processes—with mathematical optimization. People typically stop thinking further after finding answers to simple questions and give up on complex tasks after too many unsuccessful attempts. We model reasoning process as an optimization problem in “language space”, where LRMs traverse token trajectories and converge to an optimal solution. Mirroring numerical optimizers’ early‐stopping, we introduce a termination criterion that halts reasoning once reasoning converges or marginal gains fall below a preset threshold. Formally, given an input $X$ with a partial response $\mathbf{y}_{<t}$, the reasoning performance of LRM $\Pi$ is denoted as $Perf(X,\mathbf{y}_{<t}|\Pi)$. We derive a compression rule $\mathbf{c}$ to determine the stopping time $t'\in \{0,...,t-1\}$. The goal is to minimize the inference cost of $\mathbf{y}_{<t'}$ referred to as $Infer\_Cost(\mathbf{y}_{<t'})$ without degrading reasoning performance.
\begin{equation*}
\begin{aligned}
    \min_{\mathbf{c}(\cdot)} \quad  Infer\_Cost(\mathbf{y}_{<t'})~~~~\textrm{s.t.} \quad  
    &t'=\mathbf{c}(\mathbf{y}_{<t}) \\
    & Perf(X,\mathbf{y}_{<t'}|\Pi) \geq Perf(X,\mathbf{y}_{<t}|\Pi)\\ 
\end{aligned}
\end{equation*}

\textbf{Our contributions are as follows.}
\vspace{-0.8em}
\begin{itemize}
\setlength\itemsep{0pt}
  \setlength\parskip{0.2pt}
  \setlength\topsep{0pt}
  \setlength\leftmargini{0em}  
\item We propose a lightweight, training-free method for dynamic thinking compression in LRMs—mitigating both over- and underthinking—via a small verifier, enabling efficient test-time scaling without loss of reasoning accuracy.
\item We develop an asynchronous thinking compression system for industrial-scale deployment, seamlessly integrating with existing inference infrastructures.
\item Through extensive evaluations and ablations on diverse dense and Mixture of Expert (MoE) architectures (Pangu Pro MoE, \panguR, QwQ-32B, and Deepseek-R1-Distill-Qwen-32B), we demonstrate consistent reductions in reasoning cost and token usage across standard reasoning benchmarks in production settings.
\end{itemize}

\section{Related Works}
\label{sec:related_works}
\noindent\textbf{Training-based Methods.}  
Approaches in this category modify or fine-tune LRMs to generate concise reasoning. Reinforcement learning with length-based rewards prunes verbose chains~\cite{hou2025thinkprune, xia2025tokenskip, zhang2025lightthinker, team2501kimi,chen2024tencent_simpo,arora2502training}, while supervised fine-tuning teaches models to compress explanations~\cite{munkhbat2025self,yu2024distilling}. Latent-space techniques further minimize token usage by operating in a compact semantic embedding space~\cite{hao2024coconut}. Although effective, these methods demand substantial compute, risk of overfitting to specific tasks, and may degrade general-purpose capabilities.

\noindent\textbf{Self-Evaluation Methods.}  
These techniques prompt LRMs to assess their own confidence and decide when to stop reasoning. Adaptive schemes ask the model to predict the benefit of restarting~\cite{manvi2024adaptive} or to estimate certainty at key junctures~\cite{yang2025dynamic}. ST-BoN leverages embedding-distance metrics for early truncation~\cite{wang2025STBoN}. While they avoid external models, the added inference steps may introduce latency and complicate deployment.

\noindent\textbf{Model Collaboration.}  
Hybrid frameworks use auxiliary evaluators to guide decoding. Dynasor monitors semantic entropy and reward-model outputs for early stopping~\cite{fu2024dynasor,kuhn2023semantic}. Speculative Rejection uses partial-output scores from a reward model to terminate best-of-\(N\) search~\cite{sun2024speculativeRejection}. However, above works rely on the performance of the reward model, which assesses the certainty of LRMs in reaching its final answer. Another work \cite{xi2024enhancing} trains a critique model which provides step-level feedback to guide the reasoning model on the fly. It requires extra effort in dataset construction and model training. There are also concurrent works on improve LRM efficiency by generating thoughts with smaller reasoning models for speculative reasoning \cite{pan2025specreason, wang2025scot}, while they require the inefficient prefilling and decoding state switching of LRMs, LRM generation interruption, or additional fine-tuning.

\noindent Unlike prior work, our \ours{} framework is training-free and non-invasive: a lightweight verifier dynamically detects and helps truncate redundant reasoning. No extra self-evaluation steps are introduced, so our method can be easily integrated into existing inference frameworks.

\section{Method}
We propose an \textbf{efficient, verifier‐based, training‐free thinking‐compression} algorithm that dynamically prunes redundant CoT generation during online inference. Our algorithm replicates human cognitive mechanisms which utilize internal verifiers to check and stop thinking. We introduce smaller verifiers to detect redundant thinking without fine-tuning verifiers or LRMs. Designed for industrial‐scale batch processing, our framework (Fig.~\ref{fig:thinking_compression_framework}) comprises three modules: \textbf{overthinking compression} (Sec.~\ref{sec:overthinking_compression}), \textbf{underthinking compression} (Sec.~\ref{sec:underthinking_compression}), and \textbf{repetition truncation} (Sec.~\ref{sec:repetition_truncation}). Guidance prompts for halting LRM reasoning are detailed in Sec.~\ref{sec:guidance_prompt_lrm}. The online system is presented in Sec.~\ref{sec:system_design}.

\subsection{Thought Detection and Answer Extraction} 
\label{sec:thought_detection}
\vspace{-0.1cm}
The reasoning thoughts have clear structured patterns and are usually separated with \textbf{reflection tokens}, such as "\textit{\textbackslash n\textbackslash nWait}", "\textit{\textbackslash n\textbackslash nBut}", and "\textit{\textbackslash n\textbackslash nAlternatively}". In addition, LRMs normally generate answers at the end of thoughts and then verify them as in Fig.~\ref{fig:thinking_compression_framework}.

During thinking, we periodically segment the thinking process into sub-thoughts $[t_1, t_2, ..., t_{k}]$ by a method $Think\_Seg$,
\begin{equation*}
    Think\_Seg(\mathbf{y}_{<t}) = [t_1, t_2, ..., t_{k}].
\end{equation*}
Here, we separate it by reflection tokens (full list in Appendix~\ref{sec:appendix:reflection_tokens}). When new reflection tokens are detected asynchronously, we split the whole thought between two consecutive reflection tokens into sentences and concatenate the last several sentences of the current thought as the extracted \textbf{intermediate thought answers}. Formally, The last $N_{sent}$ sentences of each sub-thought $t_{l}$ are denoted as $s_l$, i.e. $s_l = Last\_sentences(t_{l}, N_{sent})$. The last several sentences are shorter but more informative than the whole sub-thought and normally contain  answers as demonstrated in Fig. ~\ref{fig:thinking_compression_framework}. We do not further process the answer but only skip the extremely short thoughts, which normally do not contain useful thoughts.

We detect predefined reflection tokens asynchronously so that LRMs continue generating tokens without throughput loss. Since reflection tokens account for only 0.5\% of all outputs and thoughts are significantly less than tokens as in Fig.~\ref{fig:count_correctness_qwq},  the verifier is invoked infrequently and can serve multiple LRMs concurrently as analyzed in Appendix \ref{sec:verifier_lrm_ratio}. It makes the overhead of redundancy checks negligible compared to the gains from thinking compression.

\vspace{-0.1cm}
\subsection{Overthinking Compression}
\label{sec:overthinking_compression}
\begin{wrapfigure}{r}{0.5\textwidth}   
  \vspace{-1em}
  \centering
  \begin{minipage}{0.5\textwidth}
\begin{algorithm}[H]
\scriptsize
\SetAlFnt{\small}
\SetAlCapFnt{\small}
\SetAlgoLined
\KwIn{Input $X$, repeat‐threshold $M$}
\KwOut{Generated output $Y$}
$stopped \gets False$;~~~~~~~~ $y_t\gets p_{\Pi}(X)$\;
$prev\_concluded\_thought \gets \mathrm{None}$;~~~~~~~~ $count \gets 0$\;
\While(\tcp*[h]{LRM iteratively generates $y_t$}){$y_{t} \neq \text{<eos>}$}{
    \tcp{Periodically segment and check for a new concluded thought}
    \tcp{$\mathcal{F}_v(p_1(s_l))$ defined in Eqn.~\ref{eqn:conclusion_detection}, $r(s_i,s_j)$ defined in Eqn.~\ref{eqn:equivalent_testing}.}
    \If{$y_t=\text{</think>}$}{$stopped=True, break;$}
    \eIf{a new segment $s_\ell$ is found and $\mathcal{F}_v(p_1(s_l))=1$ and $stopped=False$}{
        \If{$prev\_concluded\_thought \neq \mathrm{None}$}{
            \eIf{$r(prev\_concluded\_thought,\,s_{\ell})=1$}{
                $count \gets count + 1$\;
            }{
                \tcp{reset on any mismatch}
                $count \gets 0;$
            }
        }
        \If{$count \ge M$}{
            \tcp{early‐stop after $M$ repeats}
            $y_t\gets p_{\Pi}((X,\mathbf{y}_{<t}, stop\_tokens))$;\\
            $stopped\gets True, break;$
        }
        $prev\_concluded\_thought \gets s_{\ell}$\;
    }{
        $y_t\gets p_{\Pi}((X,\mathbf{y}_{<t}, y_t));$
    }
}
\Return $(\mathbf{y}_{<{t}}, y_t)$
\caption{\small{Overthinking Compression}}
\label{alg:overthinking_compression}
\end{algorithm}
\end{minipage}
\vspace{-2em}
\end{wrapfigure}

In overthinking scenarios, LRMs typically arrive at correct solutions using only $30 \sim 50\%$ of the total generated tokens, yet continue producing redundant reasoning paths or alternative justifications before finalizing an answer~\cite{chen2024tencent_simpo}.
While this may aid complex or uncertain tasks, it burdens simpler ones with uninformative content and higher latency and inference costs without accuracy gains. We propose an overthinking detection and compression algorithm that uses lightweight verifier models to prune redundant reasoning while preserving LRM accuracy. It emulates verifier confidence based human thinking termination and convergence-based early termination mechanisms of numerical optimizers.

\textbf{Simple Tasks to Utilize Smaller Models.}  We simplify overthinking detection as answer existing and equivalence checking, which are simpler binary classification problems than scoring the whole sequences.
By reducing detection to checking answer existence and comparing final outputs of consecutive intermediate thought answers, we can deploy compact (3B or 7B) verifiers with satisfactory language comprehension and instruction following capabilities without fine-tuning, greatly lowering inference overhead compared to full-sequence reward models~\cite{lightman2024lets,liu2025inference}. Moreover, training PRMs/ORMs for accurate scoring is complex and often yields unstable rewards on identical sequences, undermining reliable overthinking detection~\cite{sun2024speculativeRejection}.

\textbf{Two-Stage Verification.} To minimize intervention from answerless intermediate checks, we introduce two-stage verification. First, we confirm the candidate solution is present in the current reasoning thought. We then verify that intermediate thought answers are semantically or logically equivalent, regardless of correctness.
Both tasks can be handled by the lightweight verifier $\mathcal{F}_v$:
x

\textit{(1) Answer existence checking}: The set of thoughts with solutions $S^*$ is defined as:
\vspace{-0.1cm}
\begin{equation}
S^* = \{\,s_i\mid \mathcal{F}_v(p_{1}(s_i))\};~
\mathcal{F}_v(p_{1}(s_i))
= \mathbb{I}\Bigl[
p_{\mathcal{F}_v}(y=\text{“Yes”}\mid p_{1}(s_i))
> p_{\mathcal{F}_v}(y=\text{“No”}\mid p_{1}(s_i))
\Bigr],
\label{eqn:conclusion_detection}
\end{equation}
where $p_1(s_i)=p_1(X,s_i)$ ($X$ omitted for simplicity) is the verifier prompt for answer existence taking the problem $X$ and thought answer $s_i$ as parameters.

\textit{(2) Equivalence checking}: The equivalence $r$ between two consecutive thoughts in $S^*$ is computed as:
\vspace{-0.1cm}
\begin{equation}
r(s^*_i, s^*_{i+1}) = F_v(p_2(s^*_i,s^*_{i+1})\} = \mathbb{I}\Bigl[
p_{F_v}(y=\text{“Yes”}\mid p_2(s^*_i,s^*_{i+1}))
> p_{F_v}(y=\text{“No”}\mid p_2(s^*_i,s^*_{i+1}))
\Bigr],
\label{eqn:equivalent_testing}
\end{equation}
where $p_2(s_i, s_j)=p_2(X, s_i, s_j)$ ($X$ omitted for simplicity) is the verifier prompt for answer equivalence, taking the problem $X$ and two consecutive thought answers in $S^*$ that both achieve solutions.

\textbf{Verifier Prompts.} Verifiers take the answer existence prompt $p_1$ and answer equivalence prompt $p_2$ and directly return the binary classification results with the probabilities of "Yes" and "No" tokens. Therefore, only prefilling of verifiers is forwarded, which reduces the computational cost. The two prompts for verifiers ($p_1$ and $p_2$) in Fig.~\ref{fig:thinking_compression_framework} with placeholders are available here. Details of placeholders are in Appendix~\ref{sec:detailed_prompts}. The answer existence checking prompt $p_1$ consists of system prompt for ignoring unimportant formats and phrases, n-shot CoTs with positive and negative examples, the current question, and the intermediate answer. Similarly, the verifier only checks if two consecutive intermediate thought answers are semantically or logically equivalent with the prompt $p_2$.

\begin{minipage}[t]{0.41\textwidth}
\begin{tcolorbox}[colback=blue!4,colframe=blue!15!black,before title=\small,after title=\normalsize]
\label{prompt:promt_existence}
\scriptsize
\textbf{Verifier Prompt $p_1$ for Answer Existence:}

Check if the following sentences from a reasoning model reach an answer, 
regardless of formatting or phrasing, and solve the question. 
Return 'Yes' if the content finds a solution, otherwise 'No'. 
Return only 'Yes' or 'No' with no explanation.

\textit{\textbf{\$N-SHOT\_COTs}}~\textit{(details in Appendix)}

{Question:} \textit{\textbf{\$QUESTION}}

\textbf{Content:} \textit{\textbf{\$CONTENT}}\textbackslash n~~\text{[Yes/No]}:
\end{tcolorbox}
\end{minipage}
\hfill
\begin{minipage}[t]{0.58\textwidth}
\begin{tcolorbox}[colback=blue!5,colframe=blue!15!black,before title=\small,after title=\normalsize]
\label{prompt:promt_equivalence}
\scriptsize
\textbf{Verifier Prompt $p_2$ for Answer Equivalence}

Check if the two answers provided are logically or mathematically equivalent, regardless of formatting or phrasing. 
Return 'Yes' if they are equal in meaning/value and valid solutions to the question, otherwise 'No'. 
Return only 'Yes' or 'No' with no explanation.

\textit{\textbf{\$N-SHOT\_COTs}}~\textit{(details in Appendix)}

Question: \textit{\textbf{\$QUESTION}}

Answer 1: \textit{\textbf{\$ANSWER1}}

Answer 2: \textit{\textbf{\$ANSWER2}}

Are they equivalent? 
\text{[Yes/No]}:

\end{tcolorbox}
\end{minipage}

\textbf{Early Termination Criteria.}
We implement early termination through the prompt mechanism introduced in Sec.~\ref{sec:guidance_prompt_lrm}, specifically when the model consecutively agrees with previous reasoning steps $M$ times. The process is described in Algorithm~\ref{alg:overthinking_compression}.
We only consider thoughts in $S^*$ to bypass intermediate thoughts lacking definitive solutions.
Such a protocol analogously replicates human cognitive patterns where reasoning halts upon achieving $M+1$ consecutive consistent solutions, paralleling the convergence termination criterion in optimization algorithms.

\textbf{Resource Saving.}
By limiting inputs to 200–400 tokens instead of processing full reasoner outputs (8K–128K tokens) as did in PRMs/ORMs~\cite{sun2024speculativeRejection}, we drastically cut memory and compute overhead, boosting batch throughput and reducing verification latency with \textbf{shorter verifier prompts}.
Also, we always reuse the KV cache for the system prompt and question with \textbf{prefix caching}, then batch answers in triplets—remapping the second answer to a fixed placeholder (\texttt{\$ANSWER1} in $p_2$). This lets us retain its cache across examples, so only the cache for the other placeholder (\texttt{\$ANSWER2}, used by the first and third answers) needs updating.

\vspace{-0.5em}
\subsection{Underthinking Compression}
\label{sec:underthinking_compression}
In particularly difficult tasks, LRMs repeatedly verify intermediate steps—an indicator of uncertainty that leads to divergent, redundant reasoning. We find that when the model oscillates between different thought paths, it seldom converges: proposed solutions often fail verification, triggering further exploration that inflates latency (Fig. \ref{fig:thoughput_dicrease_over_length}) without boosting accuracy, and excessive diversity can even hinder convergence~\cite{chen2024tencent_simpo}.

To mitigate this, we repurpose the verifier’s answer‐existence/consistency feedback to detect hesitant overthinking. Concretely, if after $R_{thres}\%$ of the token budget and $N_{thres}$ reasoning rounds, the model has not produced at least three consecutive answer‐bearing chains that agree on the same solution, we flag the run as underthinking. At that point, a guidance prompt instructs the model to halt further reasoning and summarize its existing insights. It mimics human tendency to abandon difficult tasks that exceed its capabilities after multiple failed attempts.
The detailed algorithm is included in Appendix~\ref{sec:appendix:algorithm_underthinking}~Algorithm~\ref{alg:underthinking_compression}.

This mechanism relies on two principles:
(1) repeated convergence on the same solution indicates that the reasoning process is coherent and sufficiently thorough; and (2) the thresholds \(R_{thres}\) and \(N_{thres}\) must be calibrated to afford the model adequate opportunity to explore alternative solution paths before termination. 
Ablation studies on these thresholds are presented in Sec.~\ref{sec:results:underthinking_ablation}.

\vspace{-0.1cm}
\subsection{Repetition Truncation}
\label{sec:repetition_truncation}
Recurrent token loops in LRMs often persist despite probabilistic sampling tweaks (e.g., high temperature), especially when quantization is applied, and sequence length increases—trapping models in chaotic repetition. Robust repetition detection and truncation is therefore essential to prevent wasted computation and improve user experience.

Our solution uses a rolling-hash–based detector to identify repeating token-ID subsequences in real time. By updating hashes incrementally for each new token, we avoid recomputing whole-sequence hashes, enabling efficient, on-the-fly repetition checks. This dynamic algorithm is enabled by default unless otherwise specified. The effectiveness of this module is provided in Appendix~\ref{sec:results:repetition_truncation}.
\subsection{Guidance Prompts for LRMs Thinking Compression}
\label{sec:guidance_prompt_lrm}
We devised two \textbf{gentle and forceful guidance prompts} to guide terminating the reasoner’s thought process. The gentle prompt curbs overthinking by steering the model toward concluding its reasoning with \texttt{**Final Answer**\textbackslash n}, whereas the forceful prompt both prevents underthinking and breaks any repetitive loops that occur before the designated ``think end'' token (e.g. \texttt{</think>}) is emitted. In our experiments, the forceful prompt consistently halted endless generation during the reasoning stage. If repetition arises during summary generation, we immediately emit the end-of-sequence token (e.g.\ \texttt{<eos>}) to stop any further decoding generation.
\begin{tcolorbox}[
  colback=blue!5,
  colframe=blue!15!black,
  before title=\small,
  after title=\normalsize,
]
\scriptsize
\vspace{-0.2cm}
\textbf{\color{blue!15!black}Gentle Prompt (Overthinking Compression):}\\[0.5ex]
\texttt{\textbackslash n\textbackslash n I think I already thought for a long time, and I should be able to put the simple final answer after **Final Answer**\textbackslash n with \textbackslash boxed\{\} and end thinking now!}\textbf{**Final Answer**\textbackslash n}

\vspace{1ex}

\textbf{\color{blue!15!black}Forceful Prompt (Underthinking \& Repetition Truncation):}\\[0.5ex]
\texttt{\textbackslash n\textbackslash n I think I already thought for a long time, and I should be able to deduce the final answer from the content above.}
\textbf{</think>\textbackslash n\textbackslash n}
\end{tcolorbox}

\subsection{System Design}
\label{sec:system_design}

\begin{figure}
    \centering
    \includegraphics[width=1\linewidth]{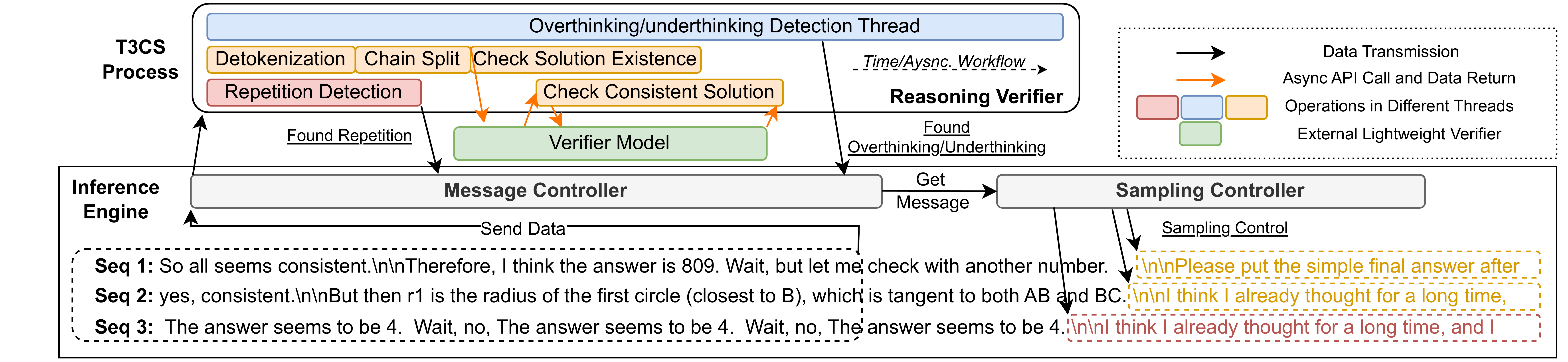}
    \vspace{-0.5cm}
    \caption{System design for the Test-Time Thinking Compression System (T3CS). The figure shows three sequences that are flagged as (1) overthinking; (2) underthinking; and (3) repetitive generation. The inference engine streams updates through the Message Controller into the external T3CS process, which issues asynchronous API calls to a lightweight verifier. The verifier detokenizes each output into reasoning chains, assesses whether they have reached valid solutions, and checks consistency across successive chains. When a sequence is identified as overthinking, underthinking, or exhibiting repetitive generation, the verifier flags it back to the Message Controller, and the Sampling Controller adjusts the logits to generate the configured tokens.}
    \label{fig:system_diagram}
\end{figure}

Our inference system is built on vLLM~\cite{kwon2023efficient} (v0.6.0) running on Ascend-910B-64GB NPUs, though it can be integrated equally well with other inference frameworks (e.g. SGLang~\cite{zheng2024sglang}) or hardware platforms (e.g. GPUs). 
As shown in Fig.~\ref{fig:system_diagram}, it comprises two tightly coupled elements: the inference engine itself, which orchestrates decoding via vLLM, and the Test-Time Thinking Compression System (T3CS), which performs real-time reasoning compression in parallel. 

As illustrated in Fig.~\ref{fig:system_diagram}, decoding proceeds in parallel across variable-length sequences via Continuous Batching. Following each forward pass, the inference engine produces logits for every active sequence; the \textbf{Message Controller} then forwards the latest token IDs to the \textbf{T3CS Process} every $N_{\text{send}}$ steps. Running asynchronously, the verifier assesses whether early convergence has occurred upon confirming valid solutions in consecutive reasoning chains, and flags any ``overthinking'' or ``underthinking'' patterns.
These flags are returned to the Message Controller, which informs the \textbf{Sampling Controller}’s adjustments to the logits, guiding the model towards particular generations. To maximize throughput, the verifier maintains a state history that prevents redundant detokenization and API calls.
In parallel, a dedicated monitoring thread watches for early-stopping conditions and issues immediate notifications through the message tunnel when they are met.

\vspace{-0.1cm}
\section{Experiments}
\vspace{-0.1cm}
\subsection{Experiment Setup}
\vspace{-0.1cm}
\textbf{Benchmarks.}
We evaluate on four commonly used reasoning benchmarks:
\textit{Mathematical benchmarks:} \textbf{MATH500}~\cite{hendrycks2021math}, \textbf{AIME24}~\cite{aime24}, \textbf{AIME25}~\cite{aime25} and \textit{Scientific benchmarks:}
\textbf{GPQA Diamond}~\cite{rein2024gpqa}. Details are available in Appendix \ref{sec:experiment_setup}.

\textbf{Metrics.}
Apart from accuracy (\textbf{Acc.}), we mainly care about the efficiency metrics. 
\textbf{Runtime} denotes the total wall-clock time to process all requests in each dataset.
\textbf{TPR} is the average Time Per Request, while \textbf{TPR-T90} is the TPR of the fastest/top 90\% requests. 
\textbf{\#Tokens(M)} is the number of generated tokens in millions. Runtime represents the total waiting time of a batch of requests, while lower TPR and TPR-T90 indicate better single user experience and higher Queries-per-Second (QPS).

\textbf{Configurations.}
All the experiments are conducted on servers with 8 Ascend 910B-64GB NPUs and 192-cores Kunpeng-920 CPUs with 1.5TB memory. 
All dataset requests (e.g., the 500 questions in MATH500)\footnote{Since AIME has only 30 questions, we replicate it eightfold to ensure the engine receives enough requests.} are submitted to vLLM concurrently, and we record key metrics such as total wall-clock time and per-request latency.
We benchmark three open-source models (Pangu Pro MoE, QwQ-32B and DeepSeek-R1-Distill-Qwen-32B(R1Q-32B)) alongside the closed-source \panguR, using fixed input/output lengths of 2K/30K tokens. Although extending outputs to $\sim$128K tokens yields marginal gains, such settings are impractical for production, so we cap the output at 30K. $M$=2, $N_{send}$=50.

\vspace{-0.1cm}
\subsection{Main Results}
\vspace{-0.1cm}
Table \ref{tab:main_results} shows that introducing thinking compression delivers consistent and substantial efficiency gains across all three models and four benchmarks (up to 70\% runtime reduction), with accuracy largely unaffected (less than 1.7\% drop).

Specifically, \textbf{runtime} is reduced by 16–39\% for QwQ-32B (e.g., from 4,413s to 3,118s on MATH500, –29.3\%), 19–39\% for \panguR (–33.8\% on MATH500), and an impressive 53–70\% for R1Q-32B (–67.0\% on MATH500; –70.0\% on GPQA Diamond). Similar reductions are seen in \textbf{TPR} (e.g., R1Q-32B’s TPR on AIME24 drops from 3,717.6s to 1,433.9s, –61.4\%). \textbf{Token usage} also drops by 8–46\% overall, with R1Q-32B showing the largest reduction (from 2.447M to 1.320M tokens, –46.1\% on GPQA Diamond).
Despite these gains, \textbf{accuracy} is preserved or even improved. QwQ-32B gains on MATH500 (+1.2\%) and AIME25 (+0.8\%), while R1Q-32B improves 2.0–13.2\% on three benchmarks (e.g., from 45.4\% to 58.6\% on GPQA Diamond). Minor regressions (e.g. –1.7\% on AIME24 for Pangu) are all under 2\%, a reasonable tradeoff for significant runtime reductions.

As demonstrated in Table \ref{tab:main_results_pangu_72b}, TrimR reduces the total runtime the Pangu Pro MoE model on all mathematical datasets (MATH500, AIME24, and AIME25) by more than 25\% with ignorable performance impact, except the AIME24 (duplicated eight times) dataset. In contrast, overthinking and underthinking are only totally triggered eight times in the GPQA Diamond dataset, because the thinking patterns of multiple-choice problems in Pangu Pro MoE are different from the question-answer (QA) counterparts. The model evaluates all options in multiple-choice problems, while it generates answers incrementally and verifies them periodically in QA problems, leading to more redundant thinking.

\begin{table}
\centering
\caption{Performance comparison of QwQ-32B, \panguR, and DeepSeek-R1-Distill-Qwen-32B on the MATH500, AIME24, AIME25, and GPQA Diamond benchmarks. Relative \textcolor{ForestGreen}{improvements} are highlighted in green, and \textcolor{red}{regressions} in red.}
\label{tab:main_results}
\resizebox{0.95\columnwidth}{!}{
\begin{tabular}{llllll}
\toprule
\textbf{Model} & \textbf{Runtime(s)} & \textbf{TPR(s)} & \textbf{TPR-T90(s)} & \textbf{Acc.} & \textbf{\#Tokens(M)} \\
\midrule
\multicolumn{6}{c}{\textbf{MATH500}} \\
\midrule
\panguR & \;\;3665 & \;\;447.4 & \;\;300.8 & 95.6\% & 1.912 \\
~~\textit{w/ TrimR}
  & \;\;2426 \smallforestgreen{-33.8\%}
  & \;\;367.6 \smallforestgreen{-17.8\%}
  & \;\;264.6 \smallforestgreen{-12.0\%}
  & 94.4\% \smallforestred{-1.2\%}
  & 1.551 \smallforestgreen{-18.9\%} \\
QwQ 32B & \;\;4413 & \;\;593.1 & \;\;439.7 & 95.6\% & 2.278  \\
~~\textit{w/ TrimR}
  & \;\;3118 \smallforestgreen{-29.3\%}
  & \;\;499.7 \smallforestgreen{-15.7\%}
  & \;\;377.4 \smallforestgreen{-14.2\%}
  & 96.8\% \smallforestgreen{1.2\%}
  & 1.953 \smallforestgreen{-14.3\%} \\
DeepSeek-R1-Distill-Qwen-32B & \;\;7602 & \;\;733.1 & \;\;278.8 & 90.4\% & 2.219\\
~~\textit{w/ TrimR}
  & \;\;2511 \smallforestgreen{-67.0\%}
  & \;\;315.8 \smallforestgreen{-56.9\%}
  & \;\;218.0 \smallforestgreen{-21.8\%}
  & 92.4\% \smallforestgreen{2.0\%}
  & 1.330 \smallforestgreen{-40.1\%} \\
\midrule
\multicolumn{6}{c}{\textbf{AIME24}} \\
\midrule
\panguR & \;\;6164 & 1912.3 & 1639.4 & 78.3\% & 2.466 \\
~~\textit{w/ TrimR}
  & \;\;3848 \smallforestgreen{-37.6\%}
  & 1299.4 \smallforestgreen{-32.0\%}
  & 1154.8 \smallforestgreen{-29.6\%}
  & 76.6\% \smallforestred{-1.7\%}
  & 2.006 \smallforestgreen{-18.6\%} \\
QwQ 32B & \;\;6992 & 2437.6 & 2138.6 & 76.6\% & 3.189  \\
~~\textit{w/ TrimR}
  & \;\;4255 \smallforestgreen{-39.1\%}
  & 1572.6 \smallforestgreen{-35.5\%}
  & 1431.6 \smallforestgreen{-33.1\%}
  & 76.6\% \smallforestgreen{-0\%}
  & 2.444 \smallforestgreen{-23.3\%} \\
DeepSeek-R1-Distill-Qwen-32B & 10299 & 3717.6 & 3156.0 & 60.0\% & 3.252 \\	
~~\textit{w/ TrimR}
  & \;\;4799 \smallforestgreen{-53.4\%}
  & 1433.9 \smallforestgreen{-61.4\%}
  & 1228.7 \smallforestgreen{-61.1\%}
  & 63.3\% \smallforestgreen{3.3\%}
  & 2.096 \smallforestgreen{-35.6\%} \\
\midrule
\multicolumn{6}{c}{\textbf{AIME25}} \\
\midrule
\panguR & \;\;9216 & 3053.4 & 2723.6 & 57.5\% & 3.117 \\
~~\textit{w/ TrimR}
  & \;\;5591 \smallforestgreen{-39.3\%}
  & 1958.6 \smallforestgreen{-35.9\%}
  & 1731.6 \smallforestgreen{-36.4\%}
  & 57.5\% \smallforestgreen{0.0\%}
  & 2.470 \smallforestgreen{-20.8\%} \\
QwQ 32B & \;\;7436 & 2771.5 & 2513.8 & 60.0\% & 3.426   \\
~~\textit{w/ TrimR}
  & \;\;6215 \smallforestgreen{-16.4\%}
  & 2302.5 \smallforestgreen{-16.9\%}
  & 2032.4 \smallforestgreen{-19.2\%}
  & 60.8\% \smallforestgreen{0.8\%}
  & 3.070 \smallforestgreen{-10.4\%} \\
DeepSeek-R1-Distill-Qwen-32B & 13055 & 5474.3 & 4861.7 & 47.9\% & 3.932 \\
~~\textit{w/ TrimR}
  & \;\;6169 \smallforestgreen{-52.7\%}
  & 1897.0 \smallforestgreen{-65.3\%}
  & 1549.6 \smallforestgreen{-68.1\%}
  & 56.3\% \smallforestgreen{8.4\%}
  & 2.434 \smallforestgreen{-38.1\%} \\
\midrule
\multicolumn{6}{c}{\textbf{GPQA Diamond}} \\
\midrule
\panguR & \;\;3120 & \;\;994.7 & \;\;866.8 & 59.1\% & 1.378 \\
~~\textit{w/ TrimR}
  & \;\;2516 \smallforestgreen{-19.4\%}
  & \;\;901.2 \smallforestgreen{-9.4\%}
  & \;\;788.9 \smallforestgreen{-9.0\%}
  & 60.1\% \smallforestgreen{1.0\%}
  & 1.273 \smallforestgreen{-7.6\%} \\
QwQ 32B & \;\;4406 & 1302.6 & 1115.0 & 66.0\% & 1.572 \\	
~~\textit{w/ TrimR}
  & \;\;3198 \smallforestgreen{-27.4\%}
  & 1170.7 \smallforestgreen{-10.1\%}
  & 1025.5 \smallforestgreen{-8.0\%}
  & 65.2\% \smallforestred{-0.8\%}
  & 1.438 \smallforestgreen{-8.5\%} \\
DeepSeek-R1-Distill-Qwen-32B & 11366 & 3568.0 & 2786.2 & 45.4\% & 2.447\\
~~\textit{w/ TrimR}
  & \;\;3411 \smallforestgreen{-70.0\%}
  & \;\;902.0 \smallforestgreen{-74.7\%}
  & \;\;720.4 \smallforestgreen{-74.1\%}
  & 58.6\% \smallforestgreen{13.2\%}
  & 1.320 \smallforestgreen{-46.1\%} \\
\bottomrule
\end{tabular}
}
\vspace{-0.6cm}
\end{table}

\begin{table}
\centering
\caption{Performance comparison of Pangu Pro MoE on the MATH500, AIME24, AIME25, and GPQA Diamond benchmarks. Relative \textcolor{ForestGreen}{improvements} are highlighted in green, and \textcolor{red}{regressions} in red.}
\label{tab:main_results_pangu_72b}
\resizebox{0.95\columnwidth}{!}{
\begin{tabular}{llllll}
\toprule
\textbf{Model} & \textbf{Runtime(s)} & \textbf{TPR(s)} & \textbf{TPR-T90(s)} & \textbf{Acc.} & \textbf{\#Tokens(M)} \\
\midrule
\multicolumn{6}{c}{\textbf{MATH500}} \\
\midrule
Pangu Pro MoE & \;\;6386 & \;\;791.3  & \;\;557.6  &  95.2\% &  1.593\\
~~\textit{w/ TrimR}
  & \;\;4801 \smallforestgreen{-24.8\%}
  & \;\;650.8 \smallforestgreen{-17.8\%}
  & \;\;478.6 \smallforestgreen{-14.2\%}
  & 94.2\% \smallforestred{-1.0\%}
  & 1.363 \smallforestgreen{-14.4\%} \\
\midrule
\multicolumn{6}{c}{\textbf{AIME24}} \\
\midrule
Pangu Pro MoE & \;\;3047 & 1263.9\;\;  & 1096.3\;\;  &  80.0\% & 0.269  
\\
~~\textit{w/ TrimR}
  & \;\;2231 \smallforestgreen{-26.8\%}
  & 1111.0 \smallforestgreen{-12.1\%}
  & \;\;989.1 \smallforestgreen{-9.8\%}
  & 76.7\% \smallforestred{-3.3\%}
  &  0.231 \smallforestgreen{-14.1\%}   
  \\
\midrule
\multicolumn{6}{c}{\textbf{AIME24 (duplicated eight times)}} \\
\midrule
Pangu Pro MoE & 10302 & 3047.2\;\;  & 2654.8\;\;  &  74.6\% & 2.141  
\\
~~\textit{w/ TrimR}
  & \;\;8200 \smallforestgreen{-20.4\%}
  & 2425.2 \smallforestgreen{-20.4\%}
  & 2073.8 \smallforestgreen{-21.9\%}
  & 69.6\% \smallforestred{-5.0\%}
  &  1.814 \smallforestgreen{-15.3\%}  
  \\
\midrule
\multicolumn{6}{c}{\textbf{AIME25}} \\
\midrule
Pangu Pro MoE & \;\;3273 & 1361.7\;\;  & 1194.4\;\;  &  70.0\% & 0.290   
\\
~~\textit{w/ TrimR}
  & \;\;2256 \smallforestgreen{-31.1\%}
  & 1225.8 \smallforestgreen{-10.0\%}
  & 1112.3 \smallforestgreen{6.9\%}
  & 66.7\% \smallforestred{-3.3\%}
  & 0.256 
  \smallforestgreen{-11.7\%} \\
\midrule
\multicolumn{6}{c}{\textbf{AIME25 (duplicated eight times)}} \\
\midrule
Pangu Pro MoE & 14528 & 4081.3\;\;  & 3602.4\;\;  &  60.8\% & 2.428 
\\
~~\textit{w/ TrimR}
  & 10631 \smallforestgreen{-26.8\%}
  & 3143.0 \smallforestgreen{-23.0\%}
  & 2729.6 \smallforestgreen{-24.2\%}
  & 60.0\% \smallforestred{-0.8\%}
  & 2.056  
  \smallforestgreen{-15.3\%} \\ 
\midrule
\multicolumn{6}{c}{\textbf{GPQA Diamond}} \\
\midrule
Pangu Pro MOE &  \;\;7780 & 1066.6  & \;\;914.7  &  76.8\% &  1.332\\
~~\textit{w/ TrimR}
  & \;\;7567 \smallforestgreen{-2.7\%}
  & 1046.3 \smallforestgreen{-1.9\%}
  & \;\;903.8 \smallforestgreen{-1.2\%}
  & 75.8\% \smallforestred{-1.0\%}
  & 1.277 \smallforestgreen{-4.1\%} \\
\bottomrule
\end{tabular}
}
\vspace{-0.6cm}
\end{table}

\subsection{Analysis}
\textbf{Effects of Compression Methods.}
Ablation results in Table \ref{tab:ablation_think_compression} demonstrate that combining overthinking and underthinking compression methods achieves the greatest efficiency gains with minimal accuracy trade-offs. For QwQ-32B on MATH500, overthinking compression alone reduces TPR by 12.0\% and tokens by 10.6\% while improving accuracy by 1.2\%, whereas underthinking compression yields smaller efficiency gains (TPR: -4.8\%, tokens: -3.3\%) with a minor 0.6\% accuracy drop. Their combination maintains the 96.8\% accuracy while achieving TPR and token reductions of 15.8\% and 14.3\%, respectively. 

For \panguR, the combined approach reduces TPR by 17.8\% and tokens by 18.9\% on MATH500 with negligible 1.2\% accuracy loss. R1Q-32B shows even stronger gains: combined compression slashes TPR by 56.9\% and tokens by 40.1\% while boosting accuracy from 90.4\% to 92.4\%.

Model-specific patterns emerge: QwQ-32B exhibits lower redundancy (overthinking TPR reduction: 12.0\% vs. 43.5\% for R1Q-32B). Conversely, R1Q-32B’s high token usage (2.447 vs. 1.572 for QwQ-32B on GPQA) reflects frequent self-verification, which dynamic compression mitigates.
Variations in compression efficacy across models and benchmarks (e.g., QwQ-32B: MATH500 TPR -12.0\% vs. AIME24 -13.1\%) underscore the need to apply both strategies to optimize efficiency across tasks.

\begin{table}
\centering
\caption{Ablation analysis of overthinking and underthinking compression, showing that both methods markedly reduce TPR and token usage without compromising reasoning accuracy.}
\label{tab:ablation_think_compression}
\resizebox{0.95\columnwidth}{!}{
\begin{tabular}{lrrrrrr}
\toprule
& \multicolumn{3}{c}{MATH500} & \multicolumn{3}{c}{AIME24} \\
\cmidrule(l){2-4}
\cmidrule(l){5-7}
& TPR & \#Tokens & Accuracy & TPR & \#Tokens & Accuracy \\
\midrule
\panguR &    -   &    -   & 95.6\% &   -    &   -    & 78.3\% \\
\textit{~w/ overthinking compression} & -16.6\% & -16.1\% & 95.4\% & -30.2\% & -15.7\% & 76.6\% \\
\textit{~w/ underthinking compression} & -10.5\% & -8.3\%  & 95.8\% & -26.0\% & -13.7\% & 75.4\% \\
\textit{~w/ both}                     & -17.8\% & -18.9\% & 94.4\% & -32.1\% & -18.6\% & 76.6\% \\
\midrule
QwQ 32B &    -   &    -   & 95.6\% &    -   &   -    & 76.6\% \\
\textit{~w/ overthinking compression} & -12.0\% & -10.6\% & 96.8\% & -13.1\% & -12.4\% & 76.6\% \\
\textit{~w/ underthinking compression} & -4.8\% & -3.3\% & 95.0\% & -27.0\% & -16.1\% & 76.3\% \\
\textit{~w/ both}                     & -15.8\% & -14.3\% & 96.8\% & -35.5\% & -23.3\% & 76.6\% \\
\midrule
DeepSeek-R1-Distill-Qwen-32B &   -    &    -   & 90.4\% &    -   &  -     & 60.0\% \\
\textit{~w/ overthinking compression} & -43.5\% & -28.1\% & 91.6\% & -49.0\% & -22.3\% & 63.3\% \\
\textit{~w/ underthinking compression} & -41.4\% & -25.0\% & 92.8\% & -57.1\% & -29.8\% & 63.8\% \\
\textit{~w/ both}                     & -56.9\% & -40.1\% & 92.4\% & -61.4\% & -35.6\% & 63.3\% \\
\bottomrule
\end{tabular}
}
\end{table}

\begin{figure}[!htp]
\centering
\begin{minipage}[b]{0.64\textwidth}
    \begin{subfigure}[b]{0.48\textwidth}
        \centering
        \includegraphics[width=\textwidth]{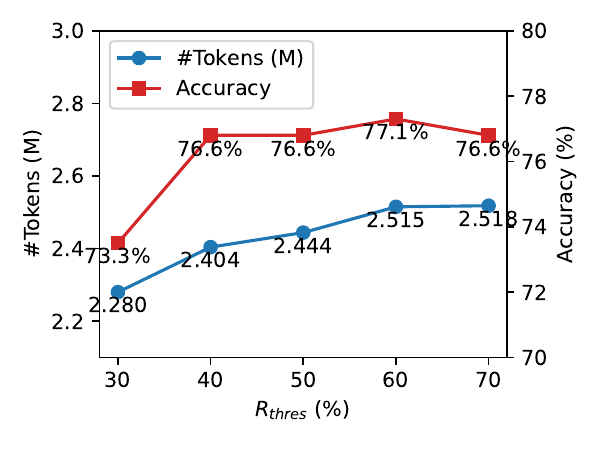}
        \vspace{-2em}
        \caption{Tokens and Acc. vs. $R_{thres}$.}
        \label{fig:r_thres}
    \end{subfigure}
    \hfill
    \begin{subfigure}[b]{0.48\textwidth}
        \centering
        \includegraphics[width=\textwidth]{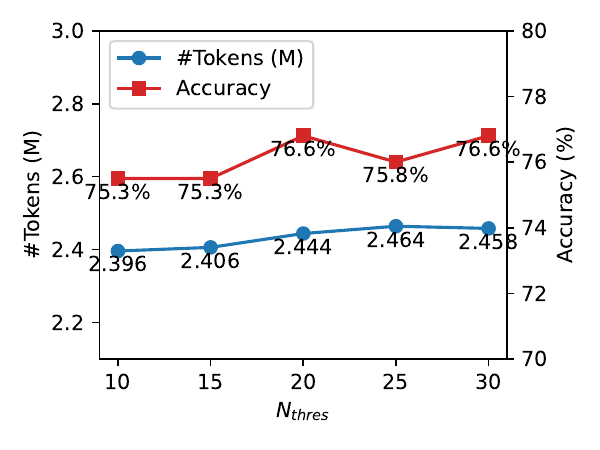}
        \vspace{-2em}
        \caption{Tokens and Acc. vs. $N_{thres}$.}
        \label{fig:n_thres}
    \end{subfigure}
    \vspace{-0.1cm}
    \caption{Effect of token‐ and round‐budget thresholds on token count and accuracy (QwQ-32B on AIME24). 
    Although performance is largely insensitive to $R_{thres}$ and $N_{thres}$, accuracy degrades when $R_{thres}<40\%$. Setting $R_{thres}=50\%$ and $N_{thres}=20$ achieves an optimal trade-off between token efficiency and accuracy.
    }
    \label{fig:underthinking_thres}
\end{minipage}
\hfill
\begin{minipage}[b]{0.35\textwidth}
    \centering
    \vspace{-0.1cm}
    \includegraphics[width=0.8\textwidth]{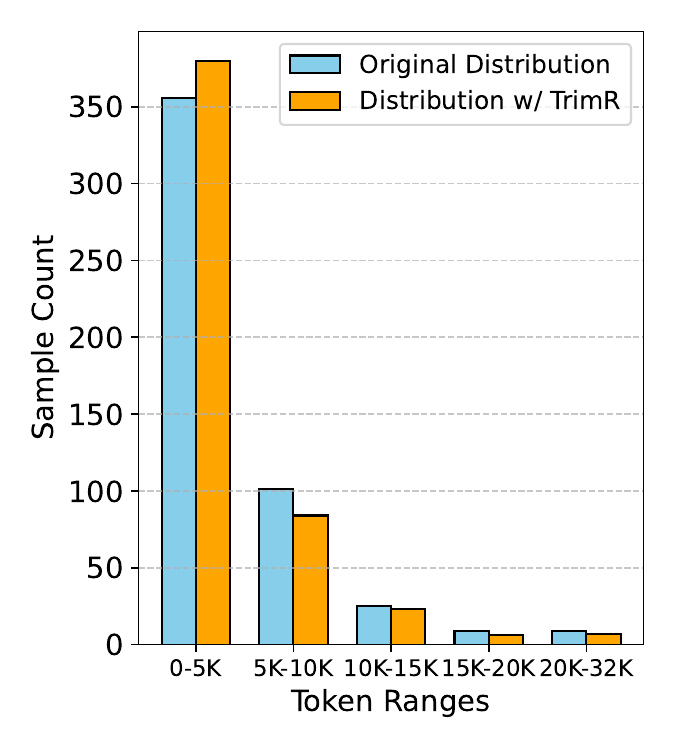}
    \vspace{-0.4cm}
    \caption{The comparison of token distributions with and without  TrimR reveals a significant leftward shift.}
    \label{fig:distribution_tokens_main}
\end{minipage}
\end{figure}

\textbf{Effects of Underthinking Thresholds.}
\label{sec:results:underthinking_ablation}
We then explore the impact of the thresholds used to flag sequences as underthinking.
Figure \ref{fig:underthinking_thres} plots accuracy and token usage against the token threshold $R_{\mathrm{thres}}$ and round threshold $N_{\mathrm{thres}}$. Raising $R_{\mathrm{thres}}$ from 30\% to 40\% increases accuracy from 73.3\% to 76.6\% while tokens rise modestly (2.28 M to 2.40 M). Beyond 50\%, accuracy plateaus but token count continues to grow. Likewise, increasing $N_{\mathrm{thres}}$ from 10 to 20 rounds boosts accuracy to 76.6\% (2.44 M tokens) with no clear gains thereafter. Accordingly, we adopt
$(R_{\mathrm{thres}}, N_{\mathrm{thres}}) = (50\%,\,20)$
as our default operating point.

\textbf{Effects on Distribution.}
Fig.~\ref{fig:distribution_tokens_main} depicts the empirical token‐count distributions for our reasoning tasks before and after applying TrimR. In the original (untrimmed) setting, approximately 64\% of problem instances fell within the lowest bin (0–5K tokens), with the remainder spread across higher token ranges (5–32K tokens). After trimming, this proportion rises to nearly 70\%, and the frequency of “long‐context” instances ($\geq$10K tokens) drops by over 25\%. In particular, the heaviest tail (20–32K tokens) is reduced by more than two‐thirds, from roughly 6\% of cases down to under 2\%.
This pronounced leftward shift in the distribution demonstrates that TrimR effectively prunes superfluous context, lowering the average token footprint per query.

\textbf{Verifier Accuracy.}
To assess chain‐level consistency, we used \panguR ~to generate full reasoning traces for all MATH500 questions, split them into T3CS‐defined chains, and manually annotated answer consistency for 684 randomly sampled adjacent chain pairs.

\begin{table}[ht]
  \centering
  \caption{Verification accuracy, and downstream performance of Pangu-7B and Qwen 2.5 Instruct verifiers on the MATH500 dataset. \emph{Verifier Acc.(\%)} denotes the fraction of correctly judged chain pairs; The rest are downstream performance on MATH500:
  \emph{MATH500 Acc.(\%)} is downstream task accuracy.}
  \label{tab:combined-verifiers}\resizebox{\textwidth}{!}{%
    \begin{tabular}{lrrrrr}
      \toprule
      Verifier                          & Verifier Acc.(\%)   & Runtime & TPR    & MATH500 Acc.(\%) & \#Tokens(M)\\
      \midrule
      Pangu-7B                          & 87.87    & 3,665    & 447.4  & 95.6           & 1.912      \\
      \quad \textit{w/o in context examples}     & 85.67    & 3,894    & 455.1  & 95.6           & 2.032      \\
      Qwen2.5-7B-Instruct                 & 86.70    & 3,722    & 459.2  & 95.0          & 1.982      \\
      \quad \textit{w/o in context examples}     & 83.48    & 3,938    & 474.2  & 94.8          & 2.103      \\
      \bottomrule
    \end{tabular}%
  }
\end{table}

As shown in Table \ref{tab:combined-verifiers}, Pangu-7B outperforms Qwen2.5-7B-Instruct in annotation accuracy (87.87\% vs. 86.70\%), speeds up thinking compression (3,665s vs. 3,722s), lowers per-request latency (447.4s vs. 459.2s), and reduces total tokens (1.912M vs. 1.982M). Downstream accuracy on MATH500 is essentially unchanged (95.6\% vs. 95.0\%), showing that occasional consistency errors have negligible effect. Omitting in-context demonstrations slightly increases runtime and token use for both verifiers (Pangu-7B: +229s, +0.120M; Qwen: +216s, +0.121M). Overall, Pangu-7B offers the best trade-off between verification fidelity and computational cost, and is selected as our default verifier.

\begin{table}[!h]
\caption{Integrating TrimR with Best-of-N (BoN) (\panguR) yields comparable efficiency improvements and token reduction while preserving the accuracy-performance trade-offs.}
\label{tab:BoN_results}
\centering
\resizebox{\textwidth}{!}{
\begin{tabular}{lllllllll}
\toprule
 & \multicolumn{4}{c}{MATH500} & \multicolumn{4}{c}{AIME24} \\
\cmidrule(l){2-5}
\cmidrule(l){6-9}
 & Runtime & TPS & Acc.(\%) & \#Tokens(M) & Runtime & TPS & Acc.(\%) & \#Tokens(M) \\
\midrule
BoN & 19,423 & 656 & 97.4 & 13.055 & 35,084 & 1,628 & 86.6 & 18.882 \\
w/ TrimR & 
16,171 \smallforestgreen{-16.7\%} & 
562 \smallforestgreen{-14.3\%} & 
97.2 \smallforestred{-0.2} & 
11.251 \smallforestgreen{-13.8\%} & 
26,911 \smallforestgreen{-23.3\%} & 
1,281 \smallforestgreen{-21.3\%} & 
83.3 \smallforestred{-3.3\%} & 
15.822 \smallforestgreen{-16.2\%} \\
\bottomrule
\end{tabular}
}
\end{table} 
\vspace{-1em}

\textbf{Think Compression in Test-time Scaling with BoN.}
Beyond sequential token extension, additional test-time scaling approaches to improve LRMs accuracy include BoN sampling~\cite{lightman2024lets}, Monte Carlo Tree Search (MCTS)~\cite{wu2025inference}, and beam search~\cite{lightman2024lets}. We integrate TrimR with BoN (N=8), as evidenced in Table~\ref{tab:BoN_results}, which demonstrates significant reductions in token consumption (-13.8-16.2\%), and runtime duration (up to 23.3\%) while maintaining performance parity (-3.3\% on AIME). 
These results highlight TrimR’s broad applicability across diverse test-time scaling frameworks.

\section{Conclusion}

This work introduces TrimR, a training-free, verifier-based framework that dynamically compresses reasoning in Large Reasoning Models (LRMs) to eliminate redundant thinking. By leveraging a lightweight pre-trained verifier to truncate unnecessary intermediate steps, TrimR significantly improves inference efficiency without compromising accuracy. Empirical results on MATH500, AIME24/25, and GPQA benchmarks demonstrate up to a 70\% reduction in runtime across diverse dense and MoE models, particularly in large-batch industrial settings. 
We also present T3CS, TrimR's online deployment system integrated with Ascend NPUs/vLLM, highlighting TrimR's scalability for high-throughput deployments. 
By balancing computational efficiency and reasoning rigor, TrimR offers a cost-effective solution for real-world LRM applications, advancing the viability of production-level AI reasoning systems.
\newpage
\bibliographystyle{plain}
\bibliography{neurips_2024}
\newpage
\appendix

\section{Algorithm of Underthinking Compression}
\label{sec:appendix:algorithm_underthinking}

We provide the formulation of underthinking compression in Algorithm~\ref{alg:underthinking_compression}.
The underthinking detection depends on the result of overthinking detection.
If a sequence can not converge to a solution within the given budget $R_{thres}$ and $N_{thres}$, we use $stop\_tokens$ (Sec.~\ref{sec:guidance_prompt_lrm}) to stop further thinking.
\begin{algorithm}[!h]
\scriptsize
\SetAlFnt{\small}
\SetAlCapFnt{\small}
\SetAlgoLined
\KwIn{Input $X$, underthinking threshold $R_{thres}$, $N_{thres}$}
\KwOut{Generated output $Y$}
$stopped \gets False$\;
$y_t\gets p_{\Pi}(X)$\;
\While(\tcp*[h]{LRM iterately generates $y_t$ in while loop}){$y_{t} \neq \text{<eos>}$}{
    \tcp{Check if this sequence is flagged as overthinking}
    $is\_overthinking \gets check~Algorithm~\ref{alg:overthinking_compression} current~state$\;
    $num\_thoughts\gets len(Think\_Seg(y_{\leq t}))$\;
    \If{$y_t=\text{</think>}$}{$stopped=True, break$}
    \eIf{$t > R_{thres}\% \cdot M$ and $num\_thoughts > N_{thres}$ and $stopped=False$}{
        $y_t\gets p_{\Pi}((X,\mathbf{y}_{<t}, stop\_tokens)),$~$stopped=True, break$\; 
    }
    {
        $y_t\gets p_{\Pi}((X,\mathbf{y}_{<t}, y_t))$
        
    }
}
\Return $(\mathbf{y}_{<{t}}, y_t)$\;
\caption{\small{Underthinking Compression}}
\label{alg:underthinking_compression}
\end{algorithm}
\vspace{-2em}
\section{More Analysis of Thinking Length and Accuracy}
\begin{figure}[h]
    \centering
    \begin{subfigure}{0.245\columnwidth}
        \centering
        \includegraphics[width=\columnwidth]{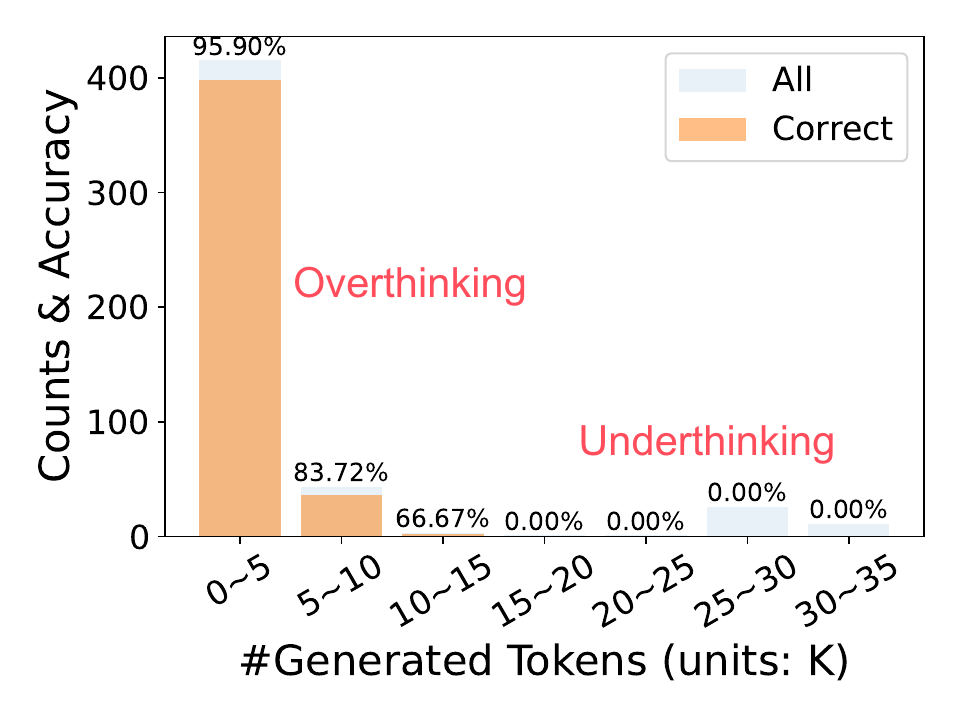}
        \vspace{-1em}
        \caption{MATH500 Accuracy}
        \label{fig:count_correctness_r1_distill_32b_math500}
    \end{subfigure}
    \begin{subfigure}{0.245\columnwidth}
        \centering
        \includegraphics[width=\columnwidth]{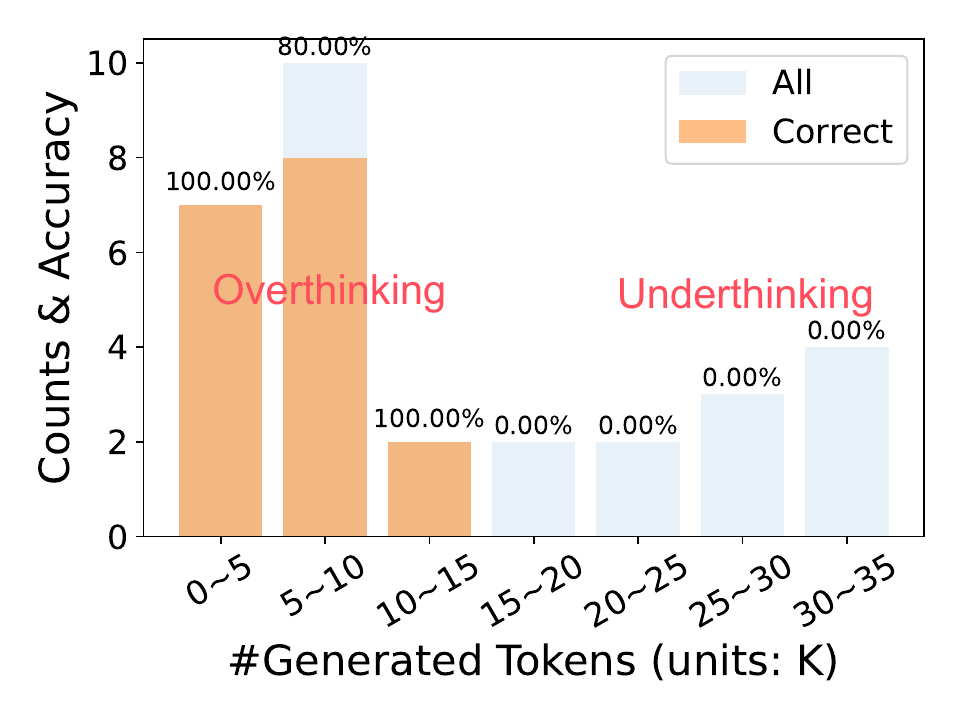}
        \vspace{-1em}
        \caption{AIME24 Accuracy}
        \label{fig:count_correctness_r1_distill_32b_aime24}
    \end{subfigure}
    \begin{subfigure}{0.245\columnwidth}
        \centering
        \includegraphics[width=\columnwidth]{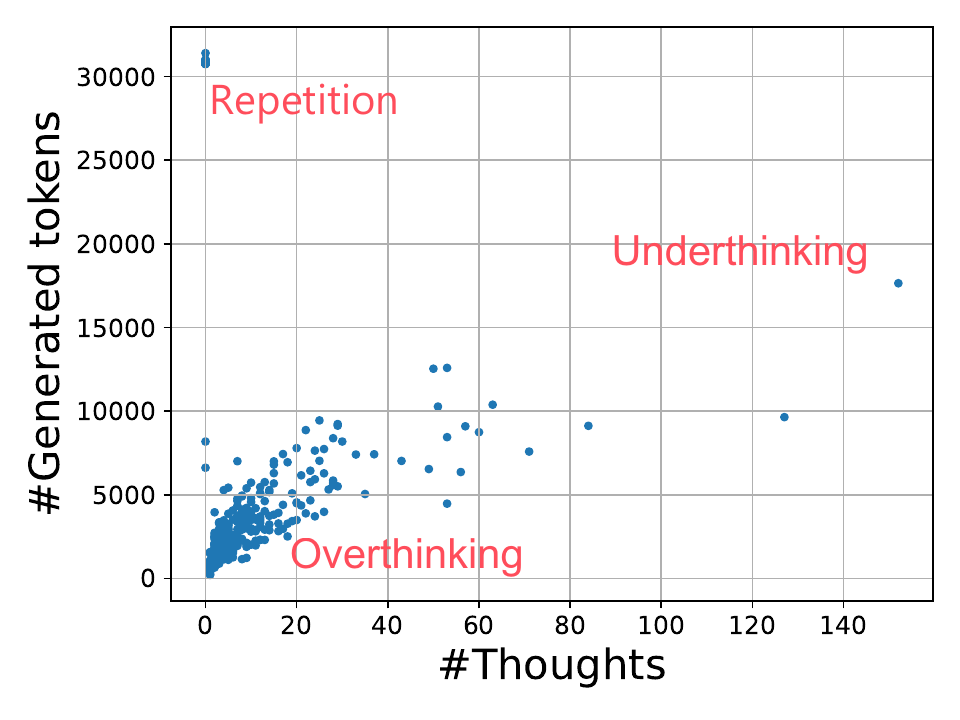}
        \vspace{-1em}
        \caption{MATH500 \#Thoughts}
        \label{fig:num_thought_tokens_r1_distill_32b_math500}
    \end{subfigure}
    \begin{subfigure}{0.245\columnwidth}
        \centering
        \includegraphics[width=\columnwidth]{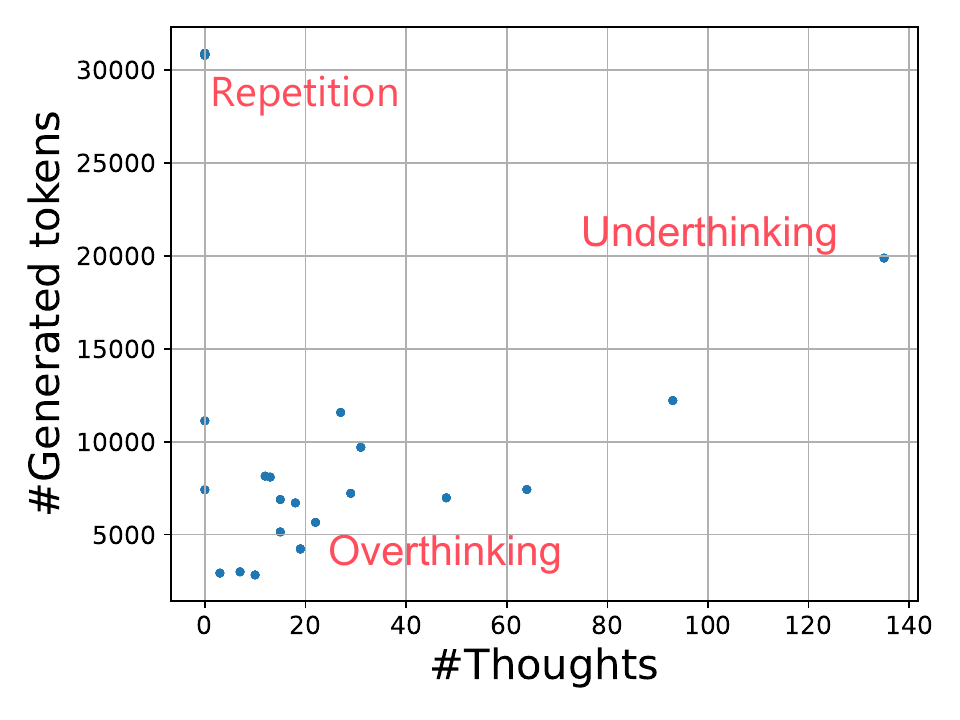}
        \vspace{-1em}
        \caption{AIME24 \#Thoughts}
        \label{fig:num_thought_tokens_r1_distill_32b_aime24}
    \end{subfigure}
    \caption{Histogram of occurrence and accuracy of decoding length and scatter plot of the number of thoughts and generated tokens of of Deepseek-R1-Distill-Qwen-32B (R1Q-32B) on MATH500 and AIME24 datasets}
    \label{fig:num_thought_tokens}
    \vspace{-2em}
\end{figure}

The distribution of decoding length occurrence and corresponding accuracy of Deepseek-R1-Distill-Qwen-32B (R1Q-32B) on the MATH500 and AIME24 datasets are available in Fig. ~\ref{fig:count_correctness_r1_distill_32b_math500}, \ref{fig:count_correctness_r1_distill_32b_aime24}. The correlation of the number of thoughts and generated tokens are available in Fig. ~\ref{fig:num_thought_tokens_r1_distill_32b_math500}, ~\ref{fig:num_thought_tokens_r1_distill_32b_aime24}.

The Deepseek-R1-Distill-Qwen-32B model is weaker than QwQ-32B and tends to generate long but wrong responses, which is underthinking. The accuracy of R1Q-32B quickly decreases from above 80\%$\sim$90\% to 0\% as visualized in Fig. ~\ref{fig:count_correctness_r1_distill_32b_math500}, \ref{fig:count_correctness_r1_distill_32b_aime24}. In contrast, QwQ-32B still is able to solve some complex questions as in Fig. ~\ref{fig:count_correctness_qwq_32b_math500}, ~\ref{fig:count_correctness_qwq_32b_aime24}.
In addition, the repetition occurs more frequently, resulting in non-stopping 32K tokens. The number of thoughts of R1Q-32B linearly correlates with the number of tokens, similar to QwQ-32B in Fig. \ref{fig:count_correctness_qwq}.

\section{Decoding throughput/latency over decoding length}
\begin{figure}
    \vspace{-0.1cm}
    \centering
    \includegraphics[width=0.4\linewidth]{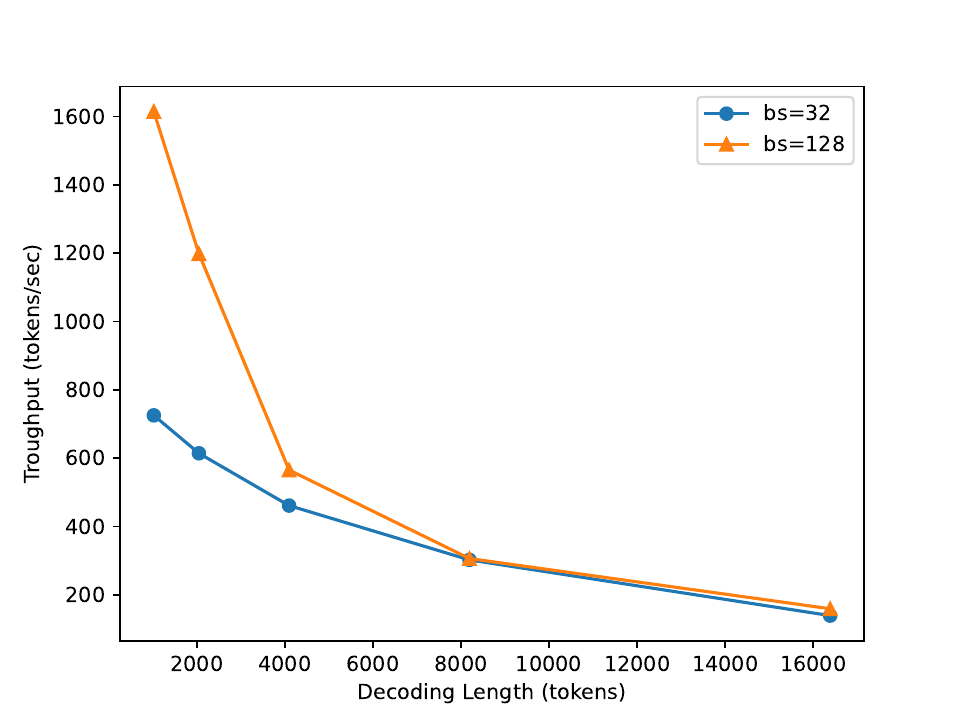}
    \caption{Throughput reduction due to increasing length of QwQ-32B.}
    \label{fig:thoughput_dicrease_over_length}
    \vspace{-0.2in}
\end{figure}
Deploying LRMs in large-scale production environments presents substantial challenges for improving reasoning efficiency.  
First, effective methods to mitigate redundant reasoning in LRMs are critical, as such inefficiencies significantly hinder the performance of inference systems in production.  
Since the generation latency of LLMs typically increases linearly with decoding length (as shown in Figure~\ref{fig:thoughput_dicrease_over_length}), reducing unnecessary token generation can yield super-linear gains in runtime reduction relative to the proportion of tokens saved. This, in turn, enhances the efficiency and scalability of test-time compute.  
Second, proposed solutions must be compatible with state-of-the-art inference infrastructures designed for large-scale deployment, such as vLLM and SGLang.

\section{Full List of Reflection Tokens}
\label{sec:appendix:reflection_tokens}

We utilize the following markers as reflection tokens to partition model reasoning into sub-thoughts:
“\textbackslash n\textbackslash nBut”,
“\textbackslash n\textbackslash nWait”,
“\textbackslash n\textbackslash nHowever”,
“\textbackslash n\textbackslash nHmm”,
“\textbackslash n\textbackslash nLet me verify this”, and 
“\textbackslash n\textbackslash nAlternatively”. 
We do not utilize those without "\textbackslash n\textbackslash n" such as "but" and "But" as reflection tokens to reduce the number of answer existence checking with verifiers, because many are in the internal step checking before approaching answers. "\textbackslash n\textbackslash n" is a strong structural separator for different thoughts in Deepseek R1 and Qwen QwQ-32B.

\section{Detailed Prompts}
\label{sec:detailed_prompts}
\begin{tcolorbox}[colback=blue!4,colframe=blue!15!black,title=Default Verifier Prompt for Answer Existence $p_1$]
\scriptsize
Check if the following sentences from a reasoning model reach a conclusion, 
regardless of formatting or phrasing, and solve the question. 
Return 'Yes' if the content finds a solution, otherwise 'No'. 
Return only 'Yes' or 'No' with no explanation.

\textbf{Example 1:}

\textbf{Question:} \( 2 + 3 = ? \)

\textbf{Content:} The answer is 5.

You should return Yes.

\textbf{Example 2:}

\textbf{Question:} \( 2 + 3 = ? \)

\textbf{Content:} I think it should be 5, but I am not sure.

You should return Yes.

\textbf{Example 3:}

\textbf{Question:} \( 2 + 3 = ? \)

\textbf{Content:} Wait, I think I made a mistake.

You should return Yes.

\textbf{Example 4:}

\textbf{Question:} If \( f(x) = \frac{3x-2}{x-2} \), what is the value of \( f(-2) + f(-1) + f(0) \)? Express your answer as a common fraction.

\textbf{Content:} 6 + 5 is 11, and 11 + 3 is 14. Yes, so \( \frac{14}{3} \). So, \( f(-2) + f(-1) + f(0) = \frac{14}{3} \).

You should return Yes.

\textbf{Example 5:}

\textbf{Question:} If \( f(x) = \frac{3x-2}{x-2} \), what is the value of \( f(-2) + f(-1) + f(0) \)? Express your answer as a common fraction.

\textbf{Content:} Since all denominators are 3, we can add the numerators: 6 + 5 + 3 = 14. Therefore, the sum is \( \frac{14}{3} \).

You should return Yes.

\textbf{Example 6:}

\textbf{Question:} If \( f(x) = \frac{3x-2}{x-2} \), what is the value of \( f(-2) + f(-1) + f(0) \)? Express your answer as a common fraction.

\textbf{Content:} Wait, another thought: When adding the fractions, is \( \frac{14}{3} \) the correct sum? Let's compute it in decimal to cross-verify. \( \frac{14}{3} \) divided is approximately 4.666...

You should return No.

\textbf{Question:} \textit{\textbf{\$QUESTION}}

\textbf{Content:} \textit{\textbf{\$CONTENT}}

\text{[Yes/No]}:

\end{tcolorbox}

\begin{tcolorbox}[colback=blue!5,colframe=blue!15!black,title=Verifier Prompt for Answer Consistency $p_2$]
\scriptsize
Check if the two answers provided are logically or mathematically equivalent, regardless of formatting or phrasing. 
Return 'Yes' if they are equal in meaning/value and a valid solution to the question, otherwise 'No'. 
Return only 'Yes' or 'No' with no explanation.

\vspace{1em}

Example 1:

Question: \( 2 + 3 = ? \)

Answer1: the answer is 5.

Answer2: the answer seems to be five.

Are they equivalent? 
\text{[Yes/No]}: \text{Yes}

\vspace{1em}

Example 2:

Question: Define 
\[
p = \sum_{k = 1}^\infty \frac{1}{k^2} \quad \text{and} \quad q = \sum_{k = 1}^\infty \frac{1}{k^3}.
\]
Find a way to write 
\[
\sum_{j = 1}^\infty \sum_{k = 1}^\infty \frac{1}{(j + k)^3}
\]
in terms of \( p \) and \( q \).

Answer 1: 
\[
(p - 1) - (q - 1) = p - 1 - q + 1 = p - q
\]
Therefore, the original double sum is equal to \( p - q \).

Answer 2: Given that the terms are decreasing and positive, the convergence seems plausible. However, since \( p - q \approx 0.4429 \) and our partial sum is about 0.3513, it's possible that the analytical result is correct. Therefore, unless there's a mistake in the analytical steps, the answer should be \( p - q \).

Are they equivalent? 
\text{[Yes/No]}: \text{Yes}

\vspace{1em}

Question: \textit{\textbf{\$QUESTION}}

Answer 1: \textit{\textbf{\$ANSWER1}}

Answer 2: \textit{\textbf{\$ANSWER2}}

Are they equivalent? 
\text{[Yes/No]}:

\end{tcolorbox}

\section{Overthinking/Underthinking examples}
\label{sec:overthinking_underthinking_examples}

The following is an example of overthinking in QwQ-32B. The question comes from AIME24. During the thinking process, reflection tokens like "\textit{\textbackslash n\textbackslash nWait}", "\textit{\textbackslash n\textbackslash nBut}" and "\textit{\textbackslash n\textbackslash nAlternatively}" appear 14 times. After the reflection tokens, the content verifies the answer "204" which first appears in the first third of the entire reasoning process.

\begin{tcolorbox}
[colback=blue!5,colframe=blue!15!black,title=An example of overthinking phenomenon (QwQ-32B)]
\small
\textbf{Question}: Every morning Aya goes for a $9$-kilometer-long walk and stops at a coffee shop afterwards. When she walks at a constant speed of $s$ kilometers per hour, the walk takes her 4 hours, including $t$ minutes spent in the coffee shop. When she walks $s+2$ kilometers per hour, the walk takes her 2 hours and 24 minutes, including $t$ minutes spent in the coffee shop. Suppose Aya walks at $s+\frac{1}{2}$ kilometers per hour. Find the number of minutes the walk takes her, including the $t$ minutes spent in the coffee shop.

\textbf{Answer}: Okay, so I have this problem about Aya's morning walk and coffee shop visit. Let me try to parse it step by step …… Convert that to minutes: 3*60 +24= 180+24=\textcolor{red}{204} minutes. Hmm, but let me confirm again. \hl{\textbackslash n\textbackslash nWait}, in the first scenario …… So is that the answer? \hl{\textbackslash n\textbackslash nWait} …… which is \textcolor{red}{204} minutes. \hl{\textbackslash n\textbackslash nBut} let me double-check …… so \textcolor{red}{204} is correct. Hmm. Alternatively, maybe I made a mistake in interpreting the total time? \hl{\textbackslash n\textbackslash nWait} …… which is \textcolor{red}{204} minutes. \hl{\textbackslash n\textbackslash nAlternatively} …… So that's consistent. \hl{\textbackslash n\textbackslash nAlternatively}, maybe there's an error in calculation …… \hl{\textbackslash n\textbackslash nWait}, hold on, let me re-derive that step …… Adding 24 minutes gives \textcolor{red}{204} minutes. So I think that's correct. \hl{\textbackslash n\textbackslash nAlternatively}, maybe the question is in minutes per kilometer or something else? …… \hl{\textbackslash n\textbackslash nAlternatively}, maybe I need to present the answer in hours converted to minutes? …… so \textcolor{red}{204} is the answer. \hl{\textbackslash n\textbackslash nWait}, but let me think again. Let me compute everything step by step once more to be thorough …… Yes, so that's exactly 180 +24=\textcolor{red}{204}. Alright, so I think that's solid …… \hl{\textbackslash n\textbackslash nAlternatively}, maybe the problem expects the answer in hours …… \hl{\textbackslash n\textbackslash nAlternatively}, maybe I made a mistake in the quadratic equation …… \hl{\textbackslash n\textbackslash nAlternatively}, maybe I should check with another approach …… \hl{\textbackslash n\textbackslash nTherefore}, I think \textcolor{red}{204} is the correct answer. Maybe I was overcomplicating, but better safe than sorry. I'll go with \textcolor{red}{204}. **Final Answer**\boxed{204} </think>

\end{tcolorbox}

The example of underthinking in QwQ-32B is present below. The question taken from AIME24 is difficult and the LRM doesn't reach the correct answer. The reasoning contents switch between different approaches frequently and can't provide an answer confidently. In fact, the figure 37/128 is made up without reasoning and the answer "475" has been checked to be wrong in previous thinking.

\begin{tcolorbox}
[colback=blue!5,colframe=blue!15!black,title=An example of underthinking phenomenon (QwQ-32B)]
\small
\textbf{Question}: Each vertex of a regular octagon is independently colored either red or blue with equal probability. The probability that the octagon can then be rotated so that all of the blue vertices end up at positions where there were originally red vertices is $\tfrac{m}{n}$, where $m$ and $n$ are relatively prime positive integers. What is $m+n$?

\textbf{Answer}: Okay, so I need to find the probability that …… \hl{\textbackslash n\textbackslash nAlternatively}, using Burnside's lemma …… \hl{\textbackslash n\textbackslash nWait}, Burnside's lemma counts the number of orbits, but I'm not sure. \hl{\textbackslash n\textbackslash nAlternatively}, let me think of it as follows …… \hl{\textbackslash n\textbackslash nTherefore}, this is the same as before. Perhaps the expected value approach …… \hl{\textbackslash n\textbackslash nBut} expectation of the union is tricky, but maybe we can approximate it …… \hl{\textbackslash n\textbackslash nAlternatively}, since B is randomly chosen, the events for different elements are not independent, but perhaps we can compute it …… \hl{\textbackslash n\textbackslash nAlternatively}, since the problem is from an Olympiad, the answer might be \textcolor{red}{37/128} or something …… \hl{\textbackslash n\textbackslash nAlternatively}, maybe the correct answer is \textcolor{red}{37/128}? \hl{\textbackslash n\textbackslash nWait}, let me think differently …… \hl{\textbackslash n\textbackslash nAlternatively}, perhaps the number is equal to the sum over all non-zero rotations of …… \hl{\textbackslash n\textbackslash nBut} without knowing the overlaps, it's hard …… This is getting too involved. Given that this is an Olympiad problem, perhaps there's a smarter way …… \hl{\textbackslash n\textbackslash nBut} without the second term, we can't compute it. \hl{\textbackslash n\textbackslash nAlternatively}, perhaps assume that the events are independent …… \hl{\textbackslash n\textbackslash nWait}, this is confusing. \hl{\textbackslash n\textbackslash nAlternatively}, if the number of subsets …… suggests that this count is too low. \hl{\textbackslash n\textbackslash nAlternatively}, perhaps the correct answer is \textcolor{red}{37/128}? \hl{\textbackslash n\textbackslash nWait} …… \hl{\textbackslash n\textbackslash nBut} the problem says "can be rotated so that all blue vertices end up at originally red vertices" …… \hl{\textbackslash n\textbackslash nTherefore}, perhaps the count is indeed \textcolor{red}{219}, giving \textcolor{red}{475}. Given that this is a competition problem, and \textcolor{red}{475} is a plausible answer (as 219+256=475), I think this is the intended answer. \hl{\textbackslash n\textbackslash nTherefore}, I will go with \textcolor{red}{475}. **Final Answer**\boxed{475} </think> 

\end{tcolorbox}

\label{sec:underthinking_tokens}

\section{Detailed Experiment Setup}
\label{sec:experiment_setup}
Benchmark description: \textbf{MATH500}~\cite{hendrycks2021math} comprises 500 problems spanning arithmetic, algebra, geometry, and calculus with varying difficulty levels. It evaluates models' ability in complex mathematical formalism, equation solving, and reasoning.  
\textbf{AIME24}~\cite{aime24} consists of Olympiad-style problems assessing logical deduction and advanced problem-solving skills. 
\textbf{AIME25}~\cite{aime25} features updated problems from the same competition as AIME24.
\textbf{GPQA Diamond}~\cite{rein2024gpqa} is a challenging dataset containing 198 multiple-choice questions, written by domain experts in biology, physics, and chemistry.

All the experiments are conducted on servers with 8 Ascend 910B-64GB NPUs and 192-cores Kunpeng-920 CPUs with 1.5TB memory. 
During decoding, TorchAir (Torch Ascend Intermediate Representation) captures the computation graph to accelerate kernel dispatch and alleviate host-bound bottlenecks.
The maximum number of concurrent decoding batches is set to 128.
We configure vLLM to pre-allocate eight scheduler steps, thereby reducing scheduling overhead. 
All dataset requests (e.g., the 500 questions in MATH500) are submitted to vLLM concurrently, and we record key metrics such as total wall-clock time and per-request latency.
We employ the Qwen2.5 math evaluation tool to score the solutions \cite{yang2024qwen2} and apply postprocessing to ensure that formatting quirks (e.g., spacing, notation style) don’t penalize valid solutions.

In the BoN experiments, we use Pangu-ORM (close-source) as the Outcome Reward Model to select the best solution from the generated N solutions.
The ratio of LRM and ORM is 1:1 in our experiments, although in production this ratio can be much higher.

\section{Effectiveness of Repetition Truncation}
\label{sec:results:repetition_truncation}

\begin{table}[htbp]
  
  \caption{Effects of Repetition Truncation over five GPQA Diamond runs.}
  \centering\label{tab:repetition_truncation}
  \centering
  \resizebox{0.92\columnwidth}{!}{
  \begin{tabular}{lccccc}
    \toprule
    Model & Runtime (H:M:S) & TPR & Accuracy & \#Tokens & Detected Repetitions \\
    \midrule
    R1Q-32B      & 4:02:51 & 5355.66 & 0.444 & 3.09M    & \textemdash         \\
    ~\textit{with repetition truncation}        & 3:09:26 & 3568.02 & 0.454 & 2.45M    & 29 out of 198      \\
    \bottomrule
  \end{tabular}
  }
\end{table}

As shown in Table \ref{tab:repetition_truncation}, applying repetition truncation to DeepSeek-R1-Distill-Qwen-32B yields substantial efficiency gains without sacrificing—and even slightly improving—accuracy:
Enabling truncation reduces total runtime by 22.0\%, cuts TPR by 50.10\%, and decreases token consumption by 20.7\%, while delivering a 1\% accuracy gain. 
Further analysis of truncated outputs confirms that most early termination occur during later stages of reasoning, where the model becomes stuck in particular attention patterns and fails to generate diverse contents. 
We also found that the guidance prompt effectively mitigates infinite repetitive outputs by steering the language model toward contextually relevant generation grounded in prior analysis.
This module thus serves as an effective, low‐overhead component of the T3CS, streamlining inference and enhancing answer clarity.

\section{Extended Analysis on Distribution}

\begin{figure}[htbp]                    
  \centering                           
  \begin{subfigure}[b]{0.48\textwidth} 
    \centering
    \includegraphics[width=\textwidth]{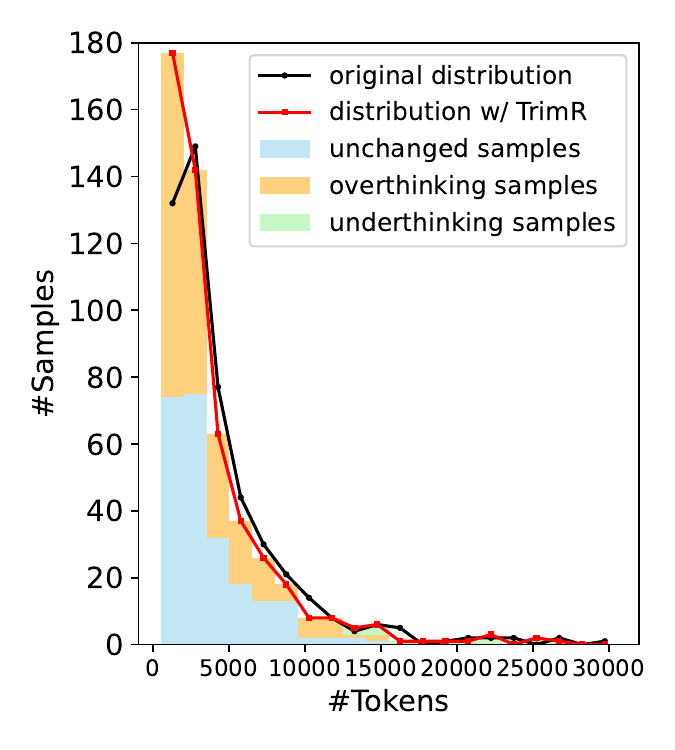} 
    \caption{The distribution of number of tokens with and without TrimR}              
    \label{fig:distribution_tokens}
  \end{subfigure}
  \hfill                                
  \begin{subfigure}[b]{0.48\textwidth} 
    \centering
    \includegraphics[width=\textwidth]{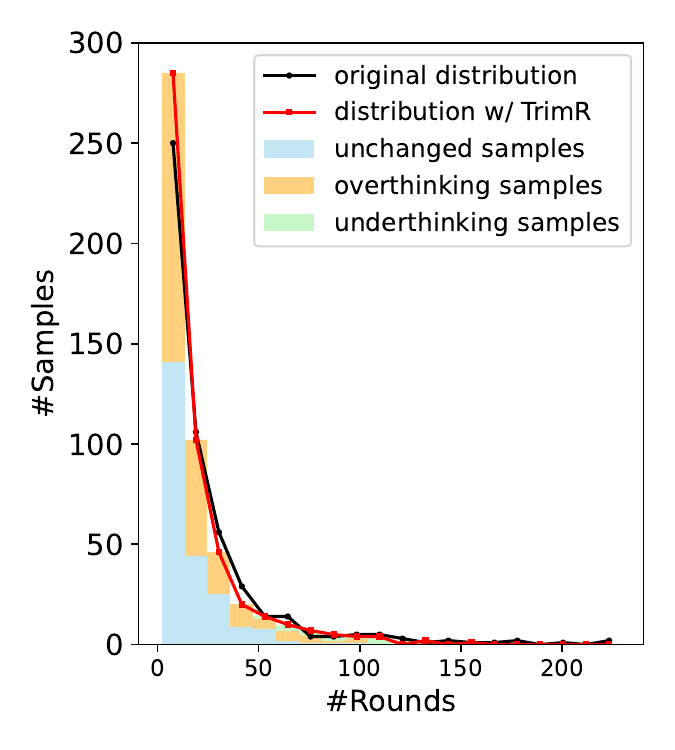}
    \caption{The distribution of number of rounds with and without TrimR}
    \label{fig:distribution_rounds}
  \end{subfigure}
  \caption{The distribution of tokens and reasoning rounds with and without TrimR (QwQ-32B on MATH500). The original distribution is indicated by the black curve, while the unchanged, overthinking, and underthinking samples are shown as stacked bars. After applying dynamic think compression, both the total tokens and number of reasoning rounds are substantially reduced compared to the original distribution.}
  \label{fig:distribution_analysis}
\end{figure}

As shown in Figure \ref{fig:distribution_tokens}, dynamic thinking compression produces a marked leftward shift in the token–usage distribution. In the uncompressed model, the ``knee'' lies around 3,500 tokens, with a substantial tail beyond 10,000 tokens. After compression, over 80\% of samples require fewer than 6,000 tokens, a roughly 30\% increase.
Also, the heavy tails (>15,000 tokens) are nearly eliminated. 
Decomposing the stacked bars reveals that unchanged samples remain tightly clustered in the low‐token bins ($\leq$4,000 tokens), overthinking samples are effectively truncated into the lower‐range bins (0–8,000 tokens), and underthinking samples (<5\% of cases) occupy lengths that are modest relative to the original tail.

A parallel effect appears in the reasoning‐round distribution (Figure 
 \ref{fig:distribution_rounds}). Prior to compression, a nontrivial fraction extends beyond 50 rounds (with outliers over 200), whereas after compression over 85\% of samples complete within 30 rounds—an increase of nearly 20 percentage points in the $\leq$20‐round regime. Overthinking cases shift from the heavy tail into the 0–30‐round interval, while underthinking cases, though rare, have been truncated to mid‐range (50–120 rounds).

Taken together, these results show that TrimR preserves valid reasoning, curtails redundant overthinking, and minimally affects cases needing additional confirmation—thus markedly improving inference efficiency in both token and round dimensions.

\section{One Verifier Can Serve Multiple LRMs}
\label{sec:verifier_lrm_ratio}
A single verifier model can simultaneously support multiple LRM instances. Runtime performance data indicates that when serving a single LRM instance, the verifier’s computational workload remains within manageable thresholds. Each LRM instance generates an average request rate of 9 requests per second, while the verifier demonstrates an average processing capacity of approximately 128 requests per second - establishing an LRM-to-verifier request ratio of 14:1 under ideal conditions. However, production systems adopt a conservative 8:1 deployment ratio to maintain operational safety margins.

When accounting for verifier infrastructure costs, TPR (Time Per Request) improvements must be adjusted by a cost-efficiency factor: $\frac{14\times 8}{14\times 8+1}=99.1\%$ (8 Ascend 910-64GB per LRM instance, 1 Ascend 910-64GB per verifier instance).
This calculation demonstrates that the performance gains per computational instance remain effectively preserved (99.1\%) despite the additional verification overhead.
It is noteworthy that enhancements in token utilization and user-perceived TPR cannot be further optimized by scaling the number of computing instances within the cluster. In contrast, \ours ~demonstrates significant performance gains by reducing user waiting times, achieving measurable improvements in latency reduction.

\end{document}